%% file: main.tex
\documentclass[10pt,twocolumn,letterpaper]{article}

\usepackage[pagenumbers]{iccv}      %

\input{preamble}

\definecolor{iccvblue}{rgb}{0.21,0.49,0.74}
\usepackage[pagebackref,breaklinks,colorlinks,allcolors=iccvblue]{hyperref}

\usepackage[capitalize]{cleveref}
\crefname{proposition}{Prop.}{Proposition}

\def\paperID{6319} %
\def\confName{ICCV}
\def\confYear{2025}

\title{Activation Subspaces for Out-of-Distribution Detection}

\newcommand{\authorstep}{\hspace{0.75cm}}
\newcommand{\affiliationstep}{\hspace{0.5cm}}

\author{
Barış Zöngür\textsuperscript{\normalfont{} 1}
\authorstep Robin Hesse\textsuperscript{\normalfont{} 1}
\authorstep Stefan Roth\textsuperscript{\normalfont{} 1,2}\\[1pt]
\small{\textsuperscript{1}TU Darmstadt \affiliationstep \textsuperscript{2}hessian.AI}
}
\begin{document}
\maketitle
\input{sec/0_abstract}  
\input{sec/1_intro}    
\input{sec/2_related_work}

\input{sec/3_method_cr}

\input{sec/4_experiments_cr}

\input{sec/5_conclusion}

{
    \small
    \bibliographystyle{ieeenat_fullname}

\input{main.bbl}
}

\input{sec/X_suppl_cr}

\end{document}

%% file: preamble.tex
\usepackage{amsmath}
\usepackage{amsthm}
\usepackage{pgfplots}
\usepackage{tabularx}
\usepackage{tikz}
\usepackage{multirow}
\usepackage{lipsum}
\usepackage{colortbl}
\usepackage{xcolor}
\usepackage{accents}
\usepackage{makecell}

\usetikzlibrary {calc,positioning,arrows.meta,graphs,shapes.misc}
\tikzset{
    nonterminal/.style={
      rectangle,
      minimum size=6mm,
      very thick,
      draw=red!50!black!50,
      top color=red!10,
      bottom color=red!10,
      },
    terminal/.style={
      rounded rectangle,
      minimum size=6mm,
      very thick,
      draw=black!50,
      top color=black!5,
      bottom color=black!5,
      }
    }

\newcolumntype{R}{@{\extracolsep{0.8cm}}r@{\extracolsep{0pt}}}%
\newcolumntype{T}{@{\extracolsep{0.5cm}}r@{\extracolsep{0pt}}}%

\usepackage{siunitx}
\newcommand{\myparagraph}[1]{\vspace{0.25em}\noindent\textbf{#1}\hspace{0.5em}}
\newcommand{\myparagraphnospace}[1]{\vspace{0.25em}\noindent\textbf{#1}}

\newtheorem{theorem}{Theorem}[section]
\newtheorem{proposition}[theorem]{Proposition}

\newcommand{\R}{\mathbb{R}}
\newcommand{\Vdirection}{\decisive{\mathbf{V}}}
\newcommand{\Vnoise}{\insignificant{\mathbf{V}}}
\newcommand{\insignificant}[1]{\accentset{\rightarrow}{#1}}
\newcommand{\decisive}[1]{\accentset{\leftarrow}{#1}}

\newcommand{\combi}[1]{\accentset{\leftrightarrow}{#1}}

%% file: sec/0_abstract.tex
\begin{abstract}
To ensure the reliability of deep models in real-world applications, out-of-distribution (OOD) detection methods aim to distinguish samples close to the training distribution (in-distribution, ID) from those farther away (OOD). 
In this work, we propose a novel OOD detection method that utilizes singular value decomposition of the weight matrix of the classification head to decompose the model's %
activations %
into \emph{decisive} and \emph{insignificant} components, which contribute maximally, respectively minimally, to the final classifier output.
We find that the subspace of \emph{insignificant} components more effectively distinguishes ID from OOD data than raw activations in regimes of large distribution shifts (Far-OOD). 
This occurs because the classification objective leaves the insignificant subspace largely unaffected, yielding features that are ``untainted'' by the target classification task.
Conversely, in regimes of smaller distribution shifts (Near-OOD), we find that activation shaping methods profit from only considering the \emph{decisive} subspace, as the \emph{insignificant} component can cause interference in the activation space. %
By combining %
two findings into a single approach, termed \emph{ActSub}, we achieve state-of-the-art results in various standard OOD benchmarks.%
\footnote{The code is available at \href{https://github.com/visinf/actsub/}{https://github.com/visinf/actsub/}.}
\end{abstract}

%% file: sec/1_intro.tex
\section{Introduction}
\label{sec:intro}

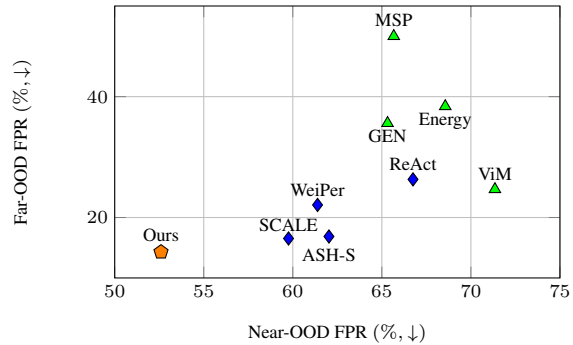
\begin{figure}[t]
\centering
\input{figures/teaser_percent}
\vspace{-0.5em}
\caption{\textit{False positive rate (FPR) for OOD detection when OOD inputs are close to (Near-OOD) and far from (Far-OOD) the ID domain.} Our approach outperforms established baselines, especially for more challenging Near-OOD examples.}
\label{fig:softmax_invariance}
\end{figure}

Improvements in the accuracy of deep neural networks (DNNs) continue to drive their deployment in a broad range of real-world settings. Consequently, ensuring the reliability of these models across a variety of conditions is becoming increasingly crucial. Since realistic applications can violate the closed-world assumption of typical DNNs, an active area of research is devoted to the development of out-of-distribution (OOD) detection methods~\cite{Yang:2024:GOD}. OOD detection is concerned with distinguishing OOD inputs, deviating from any of the predefined class distributions, from in-distribution (ID) inputs that belong to the training distribution~\cite{Hendrycks:2017:BDM}. %
Methods for OOD detection usually utilize score functions~\cite{Liu:2023:PLS, Liang:2018:ERO, Liu:2020:EOD} to output a single scalar for each input, indicating if that input is OOD or ID.
These score functions can be based on model inherent information, such as logits~\cite{Hendrycks:2022:SOD, Zhang:2023:DML, Liu:2020:EOD}, distances~\cite{Lee:2018:SUF, Sun:2022:ODD, Liu:2024:FBD, Guan:2023:RPT}, or (shaped) activations~\cite{Sun:2021:ODR, Xu:2023:VRA, Djurisic:2023:ESA, Xu:2024:STT}. %

In this work, we show that DNN activations can be decomposed into \emph{decisive directions/components}, which contribute significantly to the classifier output, and \emph{insignificant directions/components}, which have minimal effect on the model's output. Interestingly, activations still vary strongly across the insignificant subspace, which affects typical score functions for OOD detection. Based on this observation, we derive two important implications for OOD detection:
First, while decisive directions are affected by the model's classification objective --~yielding activations that specifically support the separation of the target classes (rather than ID from OOD data)~-- this effect is less pronounced for insignificant directions that are mostly ignored during training. As a result, the insignificant activation subspace is akin to powerful features from a random neural network \cite{Dirksen:2022:SCR, Hao:2022:RCS, Ulyanov:2018:DIP}, which allow to solve other classification tasks, such as OOD detection. 
While a similar proposition was already made in \cite{Cook:2020:ODN}, we extend these results by utilizing a larger set of insignificant directions. %
Second, directions from the insignificant subspace can interfere with the decisive subspace, causing different activations to produce the same output. This leads to them being handled inconsistently by activation shaping methods, which is undesirable.

To make use of these implications, we decompose the weight matrix of the linear classification head in a DNN with singular value decomposition, allowing us to decompose the weight matrix into decisive and insignificant subspaces. 
The first observation is realized by computing score functions after projecting activations into the \emph{insignificant} subspace, \ie, when decisive directions are pruned from the activations.
The second implication is addressed by applying activation shaping methods on the projection of activations into the \emph{decisive}  subspace, \ie, when interfering insignificant directions are pruned from the activations.
In a wide range of experiments, we demonstrate that the proposed activation subspaces drastically improve OOD detection, making our approach state of the art (SotA).

Specifically, our contributions are as follows: \textit{(i)} We decompose activations into orthogonal \emph{decisive} and \emph{insignificant} subspaces using singular value decomposition of the weight matrix of the classification head. %
\textit{(ii)} We show that the insignificant components provide a powerful feature space for OOD detection.
\textit{(iii)} We reduce the influence of interference between insignificant and decisive directions in the activation space by selectively applying activation shaping to the decisive subspace.  \textit{(iv)} We combine the energy score from the shaped decisive subspace and the cosine similarity score from the insignificant components. \textit{(v)} Our empirical findings suggest that when applied in combination, scores from the two activation subspaces significantly outperform previous baselines, achieving SotA results. 

%% file: figures/teaser_percent.tex
\begin{tikzpicture}
    \scriptsize
    \begin{axis}[
        xlabel={Near-OOD FPR $(\%, \downarrow)$},
        ylabel={Far-OOD FPR $(\%, \downarrow)$},
        xmin=50, xmax=75,
        ymin=10, ymax=55,
        grid=both,
        width=0.9\linewidth, %
        height=5.2cm, %
    ]
        \addplot[only marks, mark=diamond*, mark options={scale=1.25, fill=blue}] 
        coordinates {(59.76, 16.53)};
        
        \addplot[only marks, mark=diamond*, mark options={scale=1.25, fill=blue}] 
        coordinates {(61.39, 22.08)};
        
        \addplot[only marks, mark=diamond*, mark options={scale=1.25, fill=blue}] 
        coordinates {(62.03, 16.86)};

        \addplot[only marks, mark=triangle*, mark options={scale=1.25, fill=green}] 
        coordinates {(65.32, 35.61)};
        
        \addplot[only marks, mark=diamond*, mark options={scale=1.25, fill=blue}] 
        coordinates {(66.75, 26.31)};
        
        \addplot[only marks, mark=triangle*, mark options={scale=1.25, fill=green}] 
        coordinates {(68.56, 38.40)};

        \addplot[only marks, mark=triangle*, mark options={scale=1.25, fill=green}] 
        coordinates {(71.35,24.67)};

        \addplot[only marks, mark=triangle*, mark options={scale=1.25, fill=green}] 
        coordinates {(65.67,50)};

        \addplot[only marks, mark=pentagon*, mark options={scale=1.4, fill=orange}] 
        coordinates {(52.60,14.29)};
        
        \node at (axis cs:59.76, 16.7) [anchor=south] {SCALE};
        \node at (axis cs:61.39, 22.5) [anchor=south] {WeiPer};
        \node at (axis cs:62.03, 16) [anchor=north] {ASH-S};
        \node at (axis cs:65.32, 35.61) [anchor=north] {GEN};
        \node at (axis cs:66.75, 26.8) [anchor=south] {ReAct};
        \node at (axis cs:68.56, 38.40) [anchor=north] {Energy};
        \node at (axis cs:71.35, 25.1) [anchor=south] {ViM};
        \node at (axis cs:65.67, 50.5) [anchor=south] {MSP};
        \node at (axis cs:52.60, 15) [anchor=south] {Ours};

    \end{axis}
\end{tikzpicture}

%% file: sec/2_related_work.tex
\section{Related Work}
\label{sec:related}

\paragraph{Preliminaries.} 
OOD detection can be reduced to estimating a score function that separates ID and OOD inputs. A binary decision function $G(\cdot)$ determines if an input $\mathbf{x}$ is OOD ($G=1$) for model $F$ using a threshold $\tau$ and score function $S(\cdot)$ via $G(\mathbf{x},F,\tau) = [S(\mathbf{x},F)\geq \tau]$, with $[\cdot]$ denoting the Iverson bracket~\cite{Knuth:1992:TNN}.
To improve OOD detection, \textbf{training time} regularizations~\cite{HUANG:2021:TSO, Sehwag:2021:UFS, MING:2023:HEE, LU:2024:LMP, Hendrycks:2019:DAD} aim to constrain the activation space to increase ID-OOD separability. On the other hand, \textbf{post-hoc} methods utilize an already trained model and its available training data~\cite{Liu:2020:EOD, Djurisic:2023:ESA, Sun:2022:ODD, Wang:2022:ODV}. We focus on post-hoc methods, which can be further categorized by the information used to construct the score function.

\myparagraphnospace{Distance-based methods} typically rely on the activation's position in the embedding space. The distance of that position to the estimated ID distribution can be measured with a Mahalanobis distance~\cite{Lee:2018:SUF}. NuSA utilizes the norm of the projection of the activation to the null space~\cite{Cook:2020:ODN}. KNN measures the average Euclidean distance of the normalized activation~\cite{Sun:2022:ODD}. Recently, it has been shown that distances of activations to decision boundaries are discriminative for OOD detection~\cite{Liu:2024:FBD}. Alternatively, PCA-based methods use the reconstruction error as score function~\cite{Wang:2022:ODV, Guan:2023:RPT, Fang:2024:KPO}. For ID data, one expects low coverage over residual directions of PCA, as those are the directions with low expected variance. Consequently, if the norm of the residual, hence the reconstruction error, is high, the input is detected as OOD. We emphasize that the directions deemed insignificant in our work are rather distinct from PCA residuals, as our directions are determined by their contribution to the decision, not the projection's variance.

\myparagraphnospace{Logit-based methods} utilize the model's prediction instead of the activation~\cite{Hendrycks:2017:BDM, Hendrycks:2022:SOD, Liu:2020:EOD, Zhang:2023:DML}. 
For example, one can use the prediction's confidence~\cite{Hendrycks:2017:BDM} or the information entropy of the prediction~\cite{Chan:2021:EMM, Liu:2023:PLS} as the score function~\cite{Hendrycks:2017:BDM}. Similarly, the free Helmholtz energy~\cite{Liu:2020:EOD}, simply referred to as \textit{energy}, is highly used in the literature~\cite{Hofmann:2024:EHB, Lin:2021:MLO, Granz:2024:DWP, Djurisic:2023:ESA, Zhu:2022:BOD, Sun:2021:ODR}.
PCA-based methods commonly combine the reconstruction error with the energy~\cite{Wang:2022:ODV, Guan:2023:RPT}. NNGuide uses the energy as a guidance term for cosine similarity~\cite{Park:2023:NNG}. SHE uses the energy function of a Hopfield network to store class patterns~\cite{Zhang:2023:ODI}. Activation shaping methods utilize the energy of the shaped activations~\cite{Sun:2021:ODR, Xu:2023:VRA, Djurisic:2023:ESA, Xu:2024:STT, Ahn:2023:ODL, Yuan:2024:DDC}.

\myparagraphnospace{Activation shaping methods} aim to actively identify and mitigate activation channels that decrease ID-OOD separability. ReAct uses truncation to mitigate abnormal OOD activations~\cite{Sun:2021:ODR}. BATS uses inherent information in batch normalization layers~\cite{Ioffe:2015:ADN} to determine the range for truncation~\cite{Zhu:2022:BOD}. RankFeat removes the direction of the highest variance, estimated from the data matrix~\cite{Koyejo:2022:RFR}. VRA targets both abnormally high and low activations using a variational method~\cite{Xu:2023:VRA}. ASH selects top channels while pruning the rest~\cite{Djurisic:2023:ESA}, applying exponential scaling to top activations. SCALE further analyses and extends ASH, demonstrating that the key factor in ASH's effectiveness is scaling rather than pruning~\cite{Xu:2024:STT}. In parallel, methods,  such as DICE, target the weight matrix to eliminate abnormal channels~\cite{Sun:2022:LSO}. LINe uses Shapley~\cite{Shapley:1953:VPG} values to identify important neurons~\cite{Ahn:2023:ODL}. DDCS leverages the discriminability of neurons when selecting the channels to prune~\cite{Yuan:2024:DDC}. SeTAR uses low-rank approximation to refine the weight matrix, primarily targeting vision-language models~\cite{Li:2024:OSL}.

\myparagraphnospace {Gradient-based methods} investigate the gradients of the model weights. GradNorm uses the norm of the gradients~\cite{Huang:2021:OIG}. GradOrth uses the projection of gradient to the singular directions of the data matrix, following a similar motivation as PCA methods~\cite{Behpour:2023:SEO}. GAIA utilizes abnormalities in gradient-based attributions~\cite{Chen:2023:DGA}.

%% file: sec/3_method_cr.tex
\section{OOD Detection in Activation Subspaces}\label{sec:method}

\begin{figure}[t]
\centering
\input{figures/softmax_invariance}
\vspace{-0.5em}
\caption{\textit{Softmax output ($\text{softmax}_1$) for the class associated with $\text{Logit}_1$, assuming a setting with two classes.} The black lines represent contour lines for the output values \num{0.1}, \num{0.5}, and \num{0.9}. Note how they are parallel, illustrating that there is a direction in the logit space that does not affect the softmax output.}
\vspace{-1.0em}
\label{fig:softmax_invariance}
\end{figure}
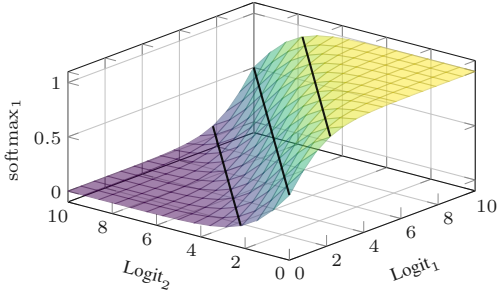

In this work, we propose a novel algorithm, termed \emph{ActSub}, for OOD detection that leverages decisive and insignificant activation subspaces of a DNN. We will first convey the intuition behind our approach
and then outline the algorithm in detail in \cref{sec:method_detail}. 

\myparagraph{Insignificant directions for OOD detection.} We let $F \colon \R^m \mapsto \R^n$ denote a DNN that maps an input $\mathbf{x} \in \R^m$ to a feature activation $\mathbf{a} \in \R^n$. To classify $\mathbf{x}$, such an activation is projected onto class logits $\mathbf{l} \in \R^c$ via a linear transformation $\mathbf{l}=\mathbf{W}\mathbf{a}$ with $\mathbf{W} \in \R^{c \times n}$ (we omit the bias $\mathbf{b}$ for simplicity of exposition and empirical insignificance). Our intuition behind using activation subspaces is based on a simple observation that was also discussed in \cite{Cook:2020:ODN,Wang:2022:ODV}: there are directions in the activation space of a DNN that have only a minimal impact on the classification output. 
With a grain of salt, we call these directions \emph{insignificant} while they turn out indeed important for OOD detection.

The existence of these directions is easy to show with the nullspace of $\mathbf{W}$ \cite{Cook:2020:ODN} --~the nullspace of a matrix $\mathbf{W}$ consists of all vectors $\mathbf{q}$ such that $\mathbf{W}\mathbf{q} = 0$~\cite{Strang:2006:LAA}. For any matrix $\mathbf{W} \in \R^{c \times n}$ with $n>c$, the nullspace is guaranteed to be non-trivial, and we let $\mathbf{Q}$ denote the orthogonal basis spanning that nullspace. For example, in an ImageNet-1k~\cite{Deng:2009:LHI, Russakovsky:2015:INS} trained ResNet-50~\cite{He:2016:DRL}, the activation $\mathbf{a}$ is in $\R^{n=2048}$ while the class logits $\mathbf{l}$ are in $\R^{c=1000}$. Since \num{2048}$>$\num{1000}, the weight matrix $\mathbf{W}$ of the model has a non-trivial nullspace and, therefore, there are directions $\mathbf{q}$ in the column space of $\mathbf{Q}$ that do not affect the model's classification output.

The existence of such insignificant directions has two interesting implications for OOD detection. First, activations along these directions still encode input statistics, making them particularly useful for OOD detection~\cite{Cook:2020:ODN}. We posit that these insignificant directions are helpful because DNNs tend to enforce activations to align with known classification directions due to the cross-entropy loss, which maximizes the class posterior~\cite{Ji:2020:DCA}. %
This presents a challenge for OOD detection because OOD activations are also biased toward high alignment with the known directions, making them difficult to distinguish from in-domain (ID) activations. Second, the insignificant directions remain ``invisible'' to the loss during training -- since the output and the gradients along these directions are constant, the model retains its behavior from the random initialization of weights. In this insignificant activation subspace, the DNN thus behaves akin to a random neural network, which are known to support challenging classification tasks \cite{Dirksen:2022:SCR}. We exploit this for OOD detection.
\begin{proposition}\label{prop:similarity}
Insignificant directions in the activation space that only marginally affect the %
output of a classifier %
can be useful for distinguishing OOD from ID activations.
\end{proposition}
\noindent
Until now, our presented intuition mostly aligns with that of \cite{Cook:2020:ODN};
however, we go beyond their work by considering a larger set of insignificant directions. %
Further, we identify a novel second implication of the existence of insignificant directions: Different activations along the insignificant directions yield the same model output, while individual activation elements $\mathbf{a}_i$ may change. Consequently, the highest activating elements can vary, potentially impacting the response of activation shaping methods \cite{Sun:2021:ODR, Xu:2023:VRA, Djurisic:2023:ESA, Xu:2024:STT}. This inconsistency can lead to different prunings for activations that result in the same model output, which is undesirable.

\begin{proposition}\label{prop:activation_shaping}
Insignificant directions in the activation space %
may interfere with decisive directions, potentially harming activation shaping methods.
\end{proposition}

\subsection{Going beyond the nullspace of weights}\label{sec:method_detail}

Now that we have established two propositions %
claiming that insignificant directions in the activation space of a DNN are helpful for OOD detection, we extend our analysis by exploring additional insignificant directions. %

\myparagraph{The softmax-invariant direction.} We assume that a classification model was trained using a cross-entropy loss, where the softmax function of the logits $\mathbf{l}$ is computed as
\begin{equation}
    \operatorname{softmax}(\mathbf{l})_i = \frac{e^{\mathbf{l}_i}}{\sum_{j=1}^c e^{\mathbf{l}_j}}\,.
\end{equation}
When plotting the softmax output for the two-dimensional case (\cf \cref{fig:softmax_invariance}), it becomes directly evident that there is a direction where the softmax output remains constant (denoted by the parallel black lines). %
It is easy to show that this direction is given by $\mathbf{1} = (1,\dots,1)^T\in\R^n$ for the general $n$-dimensional case. Due to the linearity of $\mathbf{W}\in\R^{c\times n}$, there is also a direction $\mathbf{p}$ in the activation space of the model where the softmax output remains constant when adding that direction to any activation vector $\mathbf{a}$: 
\begin{gather}
    \operatorname{softmax}(\mathbf{W}\mathbf{a})_i = \operatorname{softmax}(\mathbf{W}(\mathbf{a}+\alpha \mathbf{p}))_i \quad \forall \alpha \in \R.\label{eq:softmax-inv-direction}\raisetag{24pt}
\end{gather}
Specifically, that direction is given by $\mathbf{p} = \mathbf{W}^{\dagger}\mathbf{1}$, with $\mathbf{W}^{\dagger}$ denoting the pseudoinverse of $\mathbf{W}$.
Note that by definition, $\mathbf{p}$ is not in the null space of $\mathbf{W}$, since $\mathbf{W}\mathbf{p} = \mathbf{1} \neq 0$.

\begin{figure*}[t]
\centering
\input{figures/diagram}
\vspace{-0.5em}
\caption{\textit{Flow diagram of our proposed ActSub algorithm.} Briefly, an input $\protect\mathbf{x}$ is fed through a model $\protect{F}$ to obtain the activation $\protect\mathbf{a}$, which is decomposed into the insignificant component $\protect\insignificant{\mathbf{a}}$ and the decisive component $\protect\decisive{\mathbf{a}}$ using SVD on the weight matrix $\protect\mathbf{W}$. Our final score $\protect\combi{S}$ is a combination of $\protect\insignificant{S}$ (obtained by comparing $\protect\insignificant{\mathbf{a}}$ to insignificant activations from the training set $\protect\insignificant{\mathbf{a}}^{(i)}$) and $\protect\decisive{S}$ (obtained by applying activation shaping $\protect\Phi(\cdot)$ on $\protect\decisive{\mathbf{a}}$).
Please refer to \cref{sec:method_detail_2} for a detailed explanation.}
\vspace{-0.5em}
\label{fig:diagram}
\end{figure*}
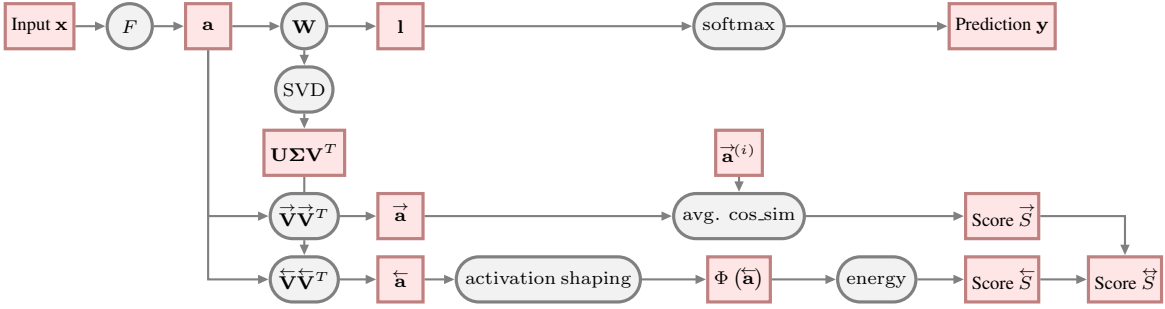

\myparagraph{Relaxing insignificant directions.} Directions in the nullspace of $\mathbf{W}$ and the above softmax-invariant direction do not change the output of the model \emph{at all} when added to any activation. However, Propositions \ref{prop:similarity} and \ref{prop:activation_shaping} also apply for insignificant directions that only \emph{marginally} affect the model output. Thus, in a more general sense, we are interested in finding all directions in the activation space that only slightly affect the classification output. To find such directions,
we make use of the singular value decomposition (SVD) by factorizing the classifier weight matrix as
\begin{equation}\label{eq:svd}
    \mathbf{W} = \mathbf{U\Sigma V}^T. 
\end{equation}
Without loss of generality, we assume that the singular values $\mathbf{\Sigma}_{ii}$ are in descending order. Intuitively, the rows in $\mathbf{V}^T$ build an orthonormal basis where the first rows correspond to the most important, \ie decisive, subspace of $\mathbf{W}$, and the last rows to the least important, \ie insignificant, subspace. The corresponding singular values indicate \emph{how} important the respective directions are.
Although the SVD in \cref{eq:svd} does not explicitly include the softmax function, we empirically find that the right singular vector corresponding to the smallest non-negative singular value almost perfectly aligns with the softmax-invariant direction, $\mathbf{p} = \mathbf{W}^{\dagger}\mathbf{1}$, described earlier, for many backbone models. We believe this property emerges during training with the softmax, as this direction is irrelevant to the loss, leading to $\mathbf{W}$ naturally ignoring it.
Further, the right singular vectors corresponding to the zero singular values lie in the nullspace of $\mathbf{W}$.
These two findings nicely connect the previously outlined theory with our actual method and further validate our approach.

With $\mathbf{W}$ being decomposed into orthonormal directions of decreasing importance, we define two auxiliary bases

\noindent\begin{minipage}{.45\linewidth}
\begin{equation*}
 \Vdirection^T=
 \begin{cases}
    \mathbf{v}^T_{i} & \text{if } i \leq k \\
    \mathbf{0}^T & \text{otherwise}
\end{cases},
\end{equation*}
\vspace{0.01em}
\end{minipage}%
\begin{minipage}{.55\linewidth}
\begin{equation}\label{eq:two_bases}
 \Vnoise^T=
 \begin{cases}
    \mathbf{0}^T & \text{if } i \leq k \\
    \mathbf{v}^T_{i} & \text{otherwise}
\end{cases},
\end{equation}
\vspace{0.01em}
\end{minipage}
for subspaces that only contain the decisive, respectively insignificant, directions of $\mathbf{V}^T$. $\mathbf{v}_{i}$ denotes the $i^\text{th}$ column in $\mathbf{V}$. %
We can now express an activation $\mathbf{a}$ in terms of these bases, effectively pruning the insignificant, respectively decisive, directions from $\mathbf{a}$:

\vspace{0.7em}
\noindent\begin{minipage}{.5\linewidth}
\begin{equation*}
 \decisive{\mathbf{a}} = \Vdirection \Vdirection^T \mathbf{a},
\end{equation*}
\vspace{0.01em}
\vspace{-0.91em}
\end{minipage}%
\begin{minipage}{.5\linewidth}
\begin{equation}\label{eq:subspace_projection}
 \insignificant{\mathbf{a}} = \Vnoise \Vnoise^T \mathbf{a}.
\end{equation}
\vspace{0.01em}
\vspace{-0.91em}
\end{minipage}
Since $\decisive{\mathbf{a}}$ is mostly relevant for the decision of the model, we call it the \textit{decisive component}. Correspondingly, we call $\insignificant{\mathbf{a}}$ the \textit{insignificant component}.

The hyperparameter $k$ in \cref{eq:two_bases} specifies the number of directions in $\mathbf{V}^T$ that are considered decisive/insignificant.
In other words, $k$ determines the degrees of freedom of the respective subspace. 
Each additional direction increases the expected norm of the projected representation; thus the norm can serve as a proxy for the amount of information captured in each subspace. 
We select $k$ such that the norms of the decisive ($\decisive{\mathbf{a}}^{(i)}$) and insignificant ($\insignificant{\mathbf{a}}^{(i)}$) components of activations $\mathbf{a}^{(i)}$ of the training samples $\mathbf{x}^{(i)}$ are as close as possible: 
\begin{equation}
  \arg\min_k \left(\frac{1}{N} \sum_{i=1}^{N} \left(\| \insignificant{\mathbf{a}}^{(i)} \| - \| \decisive{\mathbf{a}}^{(i)} \|\right) \right).
  \label{eq:ratio}
\end{equation}
Our approach eliminates the hyperparameter search on $k$, which can otherwise vastly differ for different models and numbers of classes. We provide further analysis in the supplement.

\subsection{Utilizing the activation subspaces}\label{sec:method_detail_2}

Building on the above approach to extract the decisive, respectively insignificant, directions of the activation $\mathbf{a}$, we now outline our OOD detection algorithm, combining two score functions that are based on Propositions \ref{prop:similarity} and \ref{prop:activation_shaping}.

\myparagraph{Comparing directions.} In %
Proposition~\ref{prop:similarity}, we argued that OOD and ID activations should be easier to distinguish in the insignificant subspace that only has a minor impact on the model output. %
Using the above formulation, we can directly formalize this idea with a score function $\insignificant{S}$. We extend the implementation of NNGuide~\cite{Park:2023:NNG} to insignificant components and calculate the average cosine similarity of pruned test-time input activations to pruned activations from training samples; contrary to NNGuide, we omit the guidance term. %
More specifically, we iterate over a small random subset (\num{10}\%) of the training dataset and compute the average cosine similarity $\operatorname{cos\_sim}(\cdot)$ between the insignificant component $\insignificant{\mathbf{a}}$ 
and insignificant components of the top ${N}$ samples  $\insignificant{\mathbf{a}}^{(i)}$ from the randomly sampled training subset with the highest similarity to $\insignificant{\mathbf{a}}$:
\begin{equation}
    \insignificant{S} = -\log \left(1 - \frac{1}{N} \sum_{i=1}^{N} \operatorname{cos\_sim}(\insignificant{\mathbf{a}}^{(i)}, \insignificant{\mathbf{a}})\right).
\end{equation}
While the average cosine similarity is a discriminative measure, it lies within a tight range, causing issues when combining it with our second score function $\decisive{S}$ (see below). To address this, we use $\log(1-\dots)$ on the average cosine similarity to expand the range of the values and bring both score functions into similar ranges. %

\myparagraph{Activation shaping.} Using the above formulations, the logit output of our model can be approximated as
\begin{equation}
    \mathbf{U\Sigma V}^T\mathbf{a} \approx \mathbf{U\Sigma}\Vdirection^T\decisive{\mathbf{a}}. 
\end{equation}
Recalling Proposition~\ref{prop:activation_shaping}, $\decisive{\mathbf{a}}$ only contains decisive directions, which should improve activation shaping methods $\Phi (\cdot)$ by reducing the interference from insignificant directions. We thus compute the second score function $\decisive{S}$ of our approach as %
the energy $E$~\cite{Liu:2020:EOD} over $K$ logits: 
\begin{equation}
\begin{split}
    \decisive{S} & = -E \left(  \mathbf{U\Sigma}\Vdirection^T \Phi(\decisive{\mathbf{a}}) \right) \\ & = -\log \sum_{j}^{K} \exp\left(\left(\mathbf{U\Sigma}\Vdirection^T\Phi(\decisive{\mathbf{a}})\right)_j\right).
\end{split}
\label{eq:activ-shape}
\end{equation}

\noindent\textbf{The final score function} $\combi{S}$ of our proposed \textbf{ActSub} algorithm for OOD detection is defined as
\begin{equation}\label{eq:final_score}
    \combi{S} = {\insignificant{S}}^\lambda \cdot \decisive{S},
\end{equation}
with $\lambda$ being a scalar that modulates the two score functions and is estimated via a hyperparameter search on a small range of values on the validation split (see supplement). \cref{fig:diagram} provides a flow diagram for our proposed approach.

\input{figures/pca_svd_temp}

\myparagraph{Distinction to PCA-based approaches.} %
At first glance, our proposed method may appear similar to PCA-based OOD detection methods~\cite{Wang:2022:ODV, Guan:2023:RPT, Fang:2024:KPO}. They project activations onto the directions with the lowest variance, as determined by applying PCA to the activation covariance matrix of the in-distribution training samples. However, these methods are fundamentally different from our proposed approach. Theoretically, there is no direct connection between the directions of the smallest variance in the activations and those that only marginally affect the linear classifier. However, the model could still learn to align these directions. To demonstrate that this is not the case, in \cref{fig:pca_svd}, we show the projection of the softmax-invariant direction $\mathbf{p}$ from \cref{eq:softmax-inv-direction} to the right singular vectors of the weight matrix (SVD) and the eigenvectors of the activation covariance (PCA). We use an ImageNet-1k~\cite{Deng:2009:LHI, Russakovsky:2015:INS} pre-trained ResNet-50~\cite{He:2016:DRL} with an activation dimension of $n=2048$ and samples from the ImageNet\nobreakdash-1k training split with \num{1000} classes. For SVD on the weight matrix $\mathbf W$, the softmax-invariant direction perfectly aligns with the direction of the smallest non-negative singular value. However, for PCA on the training activations, the softmax-invariant direction shows high alignment with many directions of high variance. Interestingly, the alignment is particularly high with the direction of the largest variance. %

%% file: figures/softmax_invariance.tex
\begin{tikzpicture}
    \scriptsize
    
    \begin{axis}[
        view={260+50}{35},  %
        colormap/viridis, %
        xlabel={$\text{Logit}_1$},
        ylabel={$\text{Logit}_2$},
        xlabel style={sloped}, %
        ylabel style={sloped}, %
        zlabel={$\operatorname{softmax}_1$},
        grid=major,
        xmin=0, xmax=10, ymin=0, ymax=10, %
        xtick={0,2,4,6,8,10}, ytick={0,2,4,6,8,10},
        domain=0:10, y domain=0:10, %
        opacity=0.9,
        samples=15, %
        width = 7cm,
        height = 5cm,
    ]

    \addplot3[
        surf,
        opacity=0.5
    ] 
    {exp(x)/(exp(x) + exp(y))};

    \foreach \S in {0.1} {
        \addplot3[
            thick, black, domain=0:10, y domain=0:10,
        ]
        coordinates {
            (0,  ln((exp(0)  * (1-\S)) / \S),  \S)
            (7.8,  ln((exp(7.8)  * (1-\S)) / \S),  \S)
        };
    }

    \foreach \S in {0.5} {
        \addplot3[
            thick, black, domain=0:10, y domain=0:10,
        ]
        coordinates {
            (0,  ln((exp(0)  * (1-\S)) / \S),  \S)
            (10,  ln((exp(10)  * (1-\S)) / \S),  \S)
        };
    }

    \foreach \S in {0.9} {
        \addplot3[
            thick, black, domain=0:10, y domain=0:10,
        ]
        coordinates {
            (2.2,  ln((exp(2.2)  * (1-\S)) / \S),  \S)
            (10,  ln((exp(10)  * (1-\S)) / \S),  \S)
        };
    }
    
    \end{axis}
\end{tikzpicture}

%% file: figures/diagram.tex
\begin{tikzpicture}[point/.style={circle,inner sep=0pt,minimum size=0pt,fill=black},
                    >={Stealth[inset=0pt,length=4pt,angle'=45]},thick,black!50,text=black,
                    every new ->/.style={shorten >=0pt},
                   skip loop/.style={to path={-- ++(0,#1) -| (\tikztotarget)}},
                   skip loopv/.style={to path={-- ++(#1,0) |- (\tikztotarget)}},
                    hv path/.style={to path={-| (\tikztotarget)}},
                    vh path/.style={to path={|- (\tikztotarget)}}]

  \scriptsize
  \matrix[row sep=2mm,column sep=4mm] {
    \node (ui1)   [nonterminal] {Input $\mathbf{x}$\strut}; &    \node (dot)   [terminal]    {$F$};                 &    \node (digit) [nonterminal]    {$\mathbf{a}$\strut};            &
       \node (p5)    [terminal]  {$\mathbf{W}$};                      &   \node (e)     [nonterminal]    {$\mathbf{l}$};                &
    \node (p7) [point]  {}; &    \node (softmax)     [terminal]    {$\operatorname{softmax}$\strut};                  &
    \node (p8) [point]  {}; &    \node (ui2)   [nonterminal] {Prediction $\mathbf{y}$\strut};\\
    & & & \node (SVD)     [terminal]    {$\operatorname{SVD}$}; & & & \\
    & & & \node (SVDres)     [nonterminal]    {$\mathbf{U\Sigma V}^T$}; & & & \node (trainset)     [nonterminal]    {$\insignificant{\mathbf{a}}^{(i)}$}; \\
    & & & \node (Vunimp)     [terminal]    {$\insignificant{\mathbf{V}}\insignificant{\mathbf{V}}^T$}; & \node (aunimp)     [nonterminal]    {$\insignificant{\mathbf{a}}$\strut}; & & \node (cossim)[terminal] {$\operatorname{avg.\ cos\_sim}$\strut}; & & \node (stwo) [nonterminal] {Score $\insignificant{S}$};\\
    & & & \node (Vimp)     [terminal]    {$\decisive{\mathbf{V}}\decisive{\mathbf{V}}^T$}; & \node (aimp)     [nonterminal]    {$\decisive{\mathbf{a}}$}; & \node (as)     [terminal]    {$\operatorname{activation\ shaping}$\strut}; & \node (asres) [nonterminal] {$\Phi \left( \mathbf{\decisive{a}} \right) $}; & \node (energy) [terminal] {$\operatorname{energy}$\strut}; & \node (sone) [nonterminal] {Score $\decisive{S}$}; & \node (sfinal) [nonterminal] {Score $\combi{S}$};\\
  };

  \graph {
    (ui1) -> (dot) -> (digit) -> (p5) -> (e) -> (softmax) -> (ui2);
    (p5) ->                  (SVD)   ->           (SVDres);
    (digit) -> [vh path]             (Vimp);
    (digit) -> [vh path]             (Vunimp);
    (SVDres) --             (Vunimp) -> (Vimp);
    (Vimp) -> (aimp) -> (as) -> (asres) -> (energy) -> (sone)  -> (sfinal) ;
    (Vunimp) -> (aunimp) -> (cossim) -> (stwo)  -> [hv path] (sfinal);
    (trainset) ->              (cossim);
  };
\end{tikzpicture}

%% file: figures/pca_svd_temp.tex
\begin{figure}[t]
  \centering
   \input{figures/pca_svd/pca_svd}
   \vspace{-0.5em}
   \caption{\textit{Projection of the softmax-invariant direction $\mathbf{p}$ from \cref{eq:softmax-inv-direction} onto the right singular vectors of the weight matrix (SVD) or the eigenvectors of the activation covariance (PCA).} The $x$-axis represents the respective directions in descending order of their singular values, respectively, eigenvalues. The $y$-axis shows the magnitude of the projection onto each of the corresponding directions. The softmax-invariant direction almost perfectly aligns with the right singular vector corresponding to the smallest non-negative singular value. For PCA, it aligns with many directions of high variance, especially the highest variance one.}
   \vspace{-0.5em}
   \label{fig:pca_svd}
\end{figure}
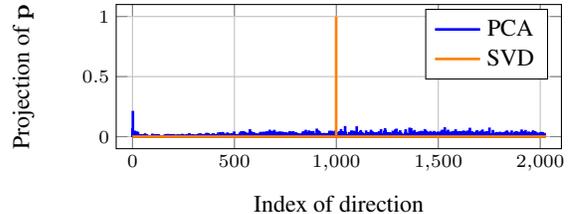

%% file: figures/pca_svd/pca_svd.tex
\begin{tikzpicture}
    \scriptsize
    \begin{axis}[
        width=0.9\linewidth,
        height=3.5cm,
        xmin=-80, xmax=2104,
        ymin=-0.1, ymax=1.1,
        xtick={0,500,1000,1500,2000},
        ytick={0,0.5,1},
        xlabel={Index of direction},
        ylabel={Projection of $\mathbf{p}$},
        xlabel style={font=\small},
        ylabel style={font=\small},
        legend pos=north east,
        legend style={font=\small},
        grid=major,
        every axis plot/.append style={line width=0.5pt}%
    ]

    \addplot[blue, ycomb, line width=1pt, forget plot] table[x index=0, y index=1] {figures/pca_svd/pca_data.txt};
    
    \addplot[orange, ycomb, line width=1pt, forget plot] table[x index=0, y index=1] {figures/pca_svd/svd_data.txt};
    
    \addplot[blue, line width=1pt] coordinates {(0,0) (2024,0)};
    \addlegendentry{PCA}

    \addplot[orange, line width=1pt] coordinates {(0,0) (2024,0)};
    \addlegendentry{SVD}

    \end{axis}
\end{tikzpicture}

%% file: sec/4_experiments_cr.tex
\section{Experiments}
\label{sec:experiments}
\input{tables/main_extension}

\myparagraph{Metrics.} We report results on two established standard metrics for OOD detection~\cite{Hendrycks:2017:BDM}: the area under the receiver operating-characteristic curve (AUC), where higher values indicate better accuracy, and the false positive rate (FPR) of the OOD inputs, where the true positive rate of ID inputs is \SI{95}{\percent}. For FPR, lower indicates better results.

\subsection{Main results}\label{subsec:main}

To demonstrate the effectiveness of our method, we follow the standard setting in the literature~\cite{Sun:2021:ODR,HUANG:2021:TSO} and report the OOD detection accuracies for an ImageNet-1k~\cite{Deng:2009:LHI, Russakovsky:2015:INS} trained ResNet-50~\cite{He:2016:DRL} for OOD data from iNaturalist~\cite{Horn:2018:ISC}, SUN~\cite{Xiao:2010:LSR}, Places~\cite{Zhou:2018:PMI}, and Textures~\cite{Cimpoi:2014:DTW} in \cref{tab:results_main_resnet50}. We use the ResNet-50 checkpoint of torchvision \cite{Paszke:2019ISH} to ensure reproducibility and fair comparison. We compare our method against established and state-of-the-art methods, focusing on different method families, such as distance-based, logit/energy-based, and activation shaping-based approaches.
To ensure fair comparisons to the respective method families, we not only report results for our final method ($\combi{S}$), but also separate it into its components: for distance-based approaches, we report OOD detection results when only considering the cosine similarities of the insignificant component ($\insignificant{S}$); for logit/energy-based approaches, we replace the decisive component (${\decisive{S}}$) of $\combi{S}$ with energy over the logits ($-E(\mathbf{W}\mathbf{a}) \cdot {\insignificant{S}}$); and for activation shaping-based approaches, we also consider only using activation shaping on the decisive component (${\decisive{S}}$). Note that method families are not strictly defined and may overlap; rather, they indicate the primary focus of each respective method. If not stated otherwise, we use the current SotA activation shaping method SCALE~\cite{Xu:2024:STT} to define $\Phi$ in ${\decisive{S}}$ (\cf Eq.\ \ref{eq:activ-shape}); however, as we show in \cref{tab:results_openood}, other activation shaping methods can be used. %

Remarkably, the different components of our method achieve superior accuracy in each respective method family. 
We achieve the highest average values among distance-based approaches, particularly excelling in detecting OOD samples from the Textures dataset. We increase the AUC by \SI{0.64}{\percent} points and reduce the FPR by \SI{6.87}{\percent} points on average compared to the closest baseline (in terms of AUC) fDBD \cite{Liu:2024:FBD}. Similarly, we outperform other methods when focusing on logit/energy-based approaches, outperforming NNGuide \cite{Park:2023:NNG} by \SI{1.05}{\percent}/\SI{2.31}{\percent} points for AUC/FPR, respectively. Our approach achieves SotA results among activation shaping methods on iNaturalist, SUN, and Places. However, the accuracy on Textures lags significantly, which impacts the overall average. When considering our final method ActSub ($\combi{S}$) and not only its individual components, we achieve SotA results on all datasets. Specifically, compared to the recent SCALE \cite{Xu:2024:STT}, our method increases the AUC by \SI{1.03}{\percent} and reduces the FPR by \SI{4.91}{\percent} points.

We hypothesize that combining both components works particularly well because the activation shaping part suffers on Textures while the distance part recovers the accuracy when combined. Both target different aspects of the distribution shift, and thus, they complement each other when combined. 
Additionally, we evaluate NNGuide~\cite{Park:2023:NNG} combined with the current SotA activation shaping-method SCALE (denoted as NN-SCALE) to demonstrate that a straightforward approach of combining SCALE with guided cosine similarity does not suffice.%

\subsection{Further backbones and datasets}\label{subsec:further_exp}
\input{tables/mobilenet}

To ensure that our findings generalize to different backbones and datasets, we next conduct a number of further experiments analogous to \cref{subsec:main}.

\myparagraph{MobileNetV2.} In \cref{tab:results_main_mobilenet}, we report results on iNaturalist, SUN, Places, and Textures for an ImageNet-1k trained MobileNetV2~\cite{Sandler:2018:IRL} (from torchvision). %
Since evaluations on this backbone are less established in related work than for ResNet-50, with fewer reported results, we include a smaller set of baselines. However, to ensure fair and complete comparisons, we reproduce key baselines, such as SCALE. Confirming the results of the ResNet-50 setup, our combined method achieves SotA results on MobileNet, with the activation shaping variant {$\accentset{\leftarrow}{S}$} lagging behind in Textures while the distance variant {$\accentset{\rightarrow}{S}$} excels on Textures.

\myparagraph{CIFAR with DenseNet-101.} Table \ref{tab:cifar_main} shows the accuracy of ActSub for CIFAR10~\cite{Krizhevsky:2009:LML} and CIFAR100~\cite{Krizhevsky:2009:LML} as ID setup. To vary the backbone even more, we use the popular DenseNet-101~\cite{Huang:2017:DCC} model, trained as in~\cite{Sun:2022:LSO,Ahn:2023:ODL}.
For OOD detection, we report the average accuracy over four datasets: SVHN~\cite{Netzer:2011:RDN}, iSUN~\cite{Xu:2015:CSW}, Places365~\cite{Zhou:2018:PMI}, and Textures~\cite{Cimpoi:2014:DTW}. Similarly to previous experiments, our method ActSub outperforms the considered baselines. We increase the AUC over SCALE by \SI{0.71}{\percent}/\SI{0.91}{\percent} points and reduce the FPR by \SI{2.47}{\percent}/\SI{4.60}{\percent} points for CIFAR10 and CIFAR100, respectively. These observations emphasize the generalization capabilities of our approach to different datasets.

\input{tables/cifar_main}

\myparagraph{OpenOOD.} We extend our experiments to the more challenging OpenOOD~\cite{Zhang:2023:EBO} benchmark. OpenOOD proposes a categorization of OOD datasets based on how close they are to the domain as Near- and Far-OOD. For ImageNet-1k as ID, OpenOOD considers NINCO~\cite{Bitterwolf:2023:FIO} and SSB-Hard~\cite{Vaze:2022:GCC} as Near-OOD; iNaturalist~\cite{Horn:2018:ISC}, Textures~\cite{Cimpoi:2014:DTW}, and OpenImage\nobreakdash-O~\cite{Wang:2022:ODV} as Far-OOD. We again use the torchvision ResNet-50 checkpoint trained on ImageNet-1k. We further evaluate ActSub in combination with ReAct, ASH-S, and SCALE, and present the results in Table \ref{tab:results_openood}. For Far-OOD, we observe a consistent improvement with our method ActSub. For ReAct, ASH-S, and SCALE, we increase the AUC by \SI{2.89}{\percent}/\SI{0.44}{\percent}/\SI{0.43}{\percent} points and reduce the FPR by \SI{10.74}{\percent}/\SI{2.64}{\percent}/\SI{2.24}{\percent} points, respectively. Furthermore, ActSub significantly outperforms SCALE and ASH\nobreakdash-S in the more challenging Near-OOD case, increasing the AUC by \SI{2.88}{\percent}/\SI{4.43}{\percent} points and reducing the FPR by \SI{7.16}{\percent}/\SI{9.57}{\percent} points, respectively. When combined with SCALE, ActSub reaches SotA results for OpenOOD.

\input{tables/openood_main}

\subsection{Ablation studies}\label{subsec:ablation}

\input{tables/pca_ablation}

Motivated by the behavior of our method on Textures --~a Far-OOD dataset compared to iNaturalist, SUN, and Places~-- we conduct ablations with the experimental setting of OpenOOD to gain a better understanding of different methods under Near- and Far-OOD settings separately.

\myparagraph{Different basis-selection strategies.}
In \cref{tab:pca_ablation_openood}, we investigate how different basis-selection strategies for decisive and insignificant subspaces affect the individual and combined components of our method. We compare our method \emph{(SVD)} against three different basis-selection strategies: \emph{(PCA)} We take principal directions from eigenvectors of the covariance matrix of activations as the decisive component; residual directions define the insignificant component. \emph{(SI-PCA)} We enforce the softmax-invariant direction to be captured in the residual subspace of PCA by setting the activation variance along the softmax-invariant direction to zero. \emph{(Null Space)} We define the insignificant component as the null space of the weight matrix, leaving the %
row space for the decisive component. For PCA and SI-PCA, we take $D=512$ directions, which corresponds to \SI{95}{\percent} of the variance.

For the cosine similarity on the insignificant component, \ie, $\accentset{\rightarrow}{S}$, we observe that PCA particularly suffers in Far-OOD. Capturing the proposed softmax-invariant direction in the PCA residuals strongly improves the accuracy, highlighting the importance of distinguishing between activation variance and activation directions that only marginally affect the classifier output. Null space improves over SI-PCA, as it targets the directions orthogonal to the model's decision. Our relaxation of the null space further improves the accuracy. 
For activation shaping on the decisive component, \ie, $\accentset{\leftarrow}{S}$, we observe similar trends as before, but now on the Near-OOD setup, with our SVD decomposition achieving the highest accuracy.
However, the Far-OOD accuracy underperforms with SVD. %
Combining information from both subspaces of our method ($\combi{S}$) again complements each other, achieving the highest accuracy in both settings.

\input{tables/component_ablation}

\myparagraph{Different components for $\accentset{\rightarrow}{S}$.} \cref{tab:part_ablation} shows the accuracy when different activation components --~\ie, the entire activation $\mathbf{a}$, the decisive component $\decisive{\mathbf{a}}$, and the insignificant component $\insignificant{\mathbf{a}}$~-- are utilized for our cosine similarity-based score $\accentset{\rightarrow}{S}$. For Far-OOD, the accuracy of $\accentset{\rightarrow}{S}$ from the decisive component $\decisive{\mathbf{a}}$ is significantly behind that of the activation $\mathbf{a}$. This observation follows our motivation on the model enforcing alignment on the directions with high significance. Supporting our approach, using the insignificant component $\insignificant{\mathbf{a}}$ yields the best results. 
The same conclusions are drawn when combining both of our score functions ($\combi{S}$).

\myparagraph{Limitations.} \textit{(i)} Similar to other distance-based approaches, our cosine similarity-based score $\accentset{\rightarrow}{S}$ relies on the training data, which limits the applicability when not available. 
\textit{(ii)} While our method is not overly sensitive to the hyperparameter $\lambda$ (see supplement), the optimal $\lambda$ can still vary for different backbones. %
\textit{(iii)} Our method relies on a linear classification head.
\textit{(iv)} Our evaluation is limited to the image classification domain, and the extension to other tasks, such as semantic segmentation, could provide valuable insights about our method's generalization.

%% file: tables/main_extension.tex
\begin{table*}
  \centering
  \scriptsize
  \begin{tabularx}{\textwidth}{@{}cX SS R SS R SS  R SS R SS@{}}
    \toprule
    &\multicolumn{1}{X}{} 
    & \multicolumn{2}{c}{\phantom{xii}\textbf{iNaturalist}} &
    & \multicolumn{2}{c}{\textbf{SUN}} &
    & \multicolumn{2}{c}{\textbf{Places}} &
    & \multicolumn{2}{c}{\textbf{Textures}} &
    & \multicolumn{2}{c}{\textbf{Average}}\\
 
\cmidrule(l{6pt}r{0pt}){3-4} \cmidrule(l{0pt}r{0pt}){6-7} \cmidrule(l{0pt}r{0pt}){9-10} \cmidrule(l{0pt}r{0pt}){12-13} \cmidrule(l{0pt}r{0pt}){15-16}

&{\textbf{Method}} 
& {AUC $(\uparrow)$} & {FPR $(\downarrow)$} &
& {AUC $(\uparrow)$} & {FPR $(\downarrow)$} & 
& {AUC $(\uparrow)$} & {FPR $(\downarrow)$} &
& {AUC $(\uparrow)$} & {FPR $(\downarrow)$} &
& {AUC $(\uparrow)$} & {FPR $(\downarrow)$} \\

\midrule

\parbox[t]{2mm}{\multirow{6}{*}{\rotatebox[origin=c]{90}{Distance}}}
&{Mahalanobis~\cite{Lee:2018:SUF}} 
& 52.65 & 97.00 &
 & 42.41 & 98.50 &
& 41.79  & 98.40 &
& 85.01  & 55.80  & 
& 55.47 & 87.43 \\

&{KNN~\cite{Wang:2022:ODV}} 
& 86.47 & 59.00 &
 & 80.72 & 68.82 &
& 75.76  & 76.28 &
& 97.07  & 11.77  & 
& 85.01  & 53.97 \\

&{ViM~\cite{Sun:2022:ODD}} 
& 87.42 & 71.85 &
 & 81.07 & 81.79 &
& 78.40  & 83.12 &
& 96.83  & 14.88  & 
& 85.93  & 62.91 \\

&{CoRP~\cite{Fang:2024:KPO}} 
& 89.32 & 50.07 &
 & 83.74 & 62.56 &
& 78.91  & 72.76 &
& 98.14  & 9.02  & 
& 87.53  & 48.60 \\

&{fDBD~\cite{Liu:2024:FBD}} 
& 93.67 & 40.24 &
 & 86.97 & 60.60 &
& 84.27 & 66.40  &
& 92.12 & 37.50  & 
& 89.26 & 51.19 \\

\cmidrule(l{6pt}r{0pt}){2-16}

&{$\accentset{\rightarrow}{S}$} (ours)
& 91.77 & 42.80 &
& 87.43 & 56.71 &
& 82.08 & 69.23  &
& \bfseries98.33 & \bfseries8.53  & 
& 89.90 & 44.32 \\

\midrule
\parbox[t]{2mm}{\multirow{9}{*}{\rotatebox[origin=c]{90}{Logit}}}
&{MSP~\cite{Hendrycks:2017:BDM}} 
& 88.17 & 51.44 &
& 79.95 & 72.04 &
& 78.84 & 74.34 &
& 54.90 & 78.69 &  
& 81.41 & 63.18 \\

&{ODIN~\cite{Liang:2018:ERO}} 
& 91.32 & 41.07 & 
& 84.71 & 64.63 & 
& 81.95 & 68.36 & 
& 85.77 & 50.55 & 
& 85.94 & 56.15 \\

&{Energy~\cite{Liu:2020:EOD}} 
& 89.95 & 55.72 &
& 85.89 & 59.26 &
& 82.86 & 64.92 &
& 85.99 & 53.72 &  
& 86.17 & 58.41\\

&{NNGuide~\cite{Park:2023:NNG}} 
& 95.12 & 25.73 &
& 91.21 & 37.18 &
& 88.67 & 46.97 &
& 92.30 & 27.70 & 
& 91.82 & 34.39\\

&{NRE~\cite{Guan:2023:RPT}} 
& 91.09 & 50.36 &
& 87.55 & 54.19 &
& 84.00 & 64.13 &
& 92.59  & 29.33 & 
&  88.81 & 49.50 \\

&{DML~\cite{Zhang:2023:DML}} 
& 91.61 & 47.32 &
& 86.14  & 57.40 &
& 84.68 & 61.43 &
& 86.72 & 52.80 & 
& 87.28 & 54.74 \\

&{SHE~\cite{Zhang:2023:ODI}} 
& 90.18 & 34.22 &
& 84.69  & 54.19 &
& 90.15 & 45.35 &
& 87.93 & 45.09 & 
& 88.24 & 44.71 \\

&{CoRP~\cite{Fang:2024:KPO}} 
& 95.15 & 26.85 &
& 90.76 & 40.38 &
& 87.35 & 51.26 &
& 97.17  & 12.11 & 
&  92.61 & 32.65 \\

\cmidrule(l{6pt}r{0pt}){2-16}
&{$-E \cdot \accentset{\rightarrow}{S}$} (ours)
& 95.21	 & 26.30 &
 & 91.20 & 39.82 &
& 87.63	& 51.84	 &
& 97.43 & 10.35  & 
& 92.87 & 32.08 \\ 
       
\midrule

\parbox[t]{2mm}{\multirow{14}{*}{\rotatebox[origin=c]{90}{Activation shaping}}}

&{ReAct~\cite{Sun:2021:ODR}} 
& 96.22 & 20.38 &
& 94.20 & 24.20 &
& 91.58 & 33.85 &
& 89.80 & 47.30 & 
& 92.95 & 31.43\\

&{BATS~\cite{Zhu:2022:BOD}} 
& 97.67 & 12.57 &
& 95.33 & 22.62 &
& 91.83 & 34.34 &
& 92.27 & 38.90 & 
& 94.28 & 27.11\\

&{DICE~\cite{Sun:2022:LSO}} 
& 96.24 & 18.64 &
& 93.94 & 25.45 &
& 90.67 & 36.86 &
& 92.74 & 28.07 & 
& 93.40 & 27.25\\

&{NRE~\cite{Guan:2023:RPT}} 
& 97.97 & 10.17 &
& 95.80 & 18.50 &
& 93.39 & 27.31 &
& 95.95 & 18.67 & 
& 95.76 & 18.66\\

&{VRA~\cite{Xu:2023:VRA}} 
& 97.12 & 15.70 &
& 94.25 & 26.94 &
& 91.27 & 37.85 &
& 95.62 & 21.47 & 
& 94.57 & 25.49\\

&{LINe~\cite{Ahn:2023:ODL}} 
& 97.56  & 12.26 &
& 95.26 & 19.48 &
& 92.85 & 28.52 &
& 94.44 & 22.54 & 
& 95.03 & 20.70\\

&{ASH-S~\cite{Djurisic:2023:ESA}} 
& 97.87 & 11.49 &
& 94.02 & 27.98 &
& 90.98 & 39.78 &
& 97.60 & 11.93 & 
& 95.12 & 22.80\\

&{CoRP~\cite{Fang:2024:KPO}} 
& 97.85 & 10.77 &
& 95.75 & 18.70 &
& 93.13 & 28.69 &
& 97.21 & 12.57 &
& 95.98 & 17.68 \\
 
&{fDBD~\cite{Liu:2024:FBD}} 
& 98.07 & 10.19 &
& 94.87 & 24.58 &
& 92.00 & 36.12 &
& 97.48 & 12.52 &
& 95.61 & 20.85 \\

&{DDCS~\cite{Yuan:2024:DDC}} 
& 97.85 & 11.63 &
& 95.68 & 18.63 &
& 92.89 & 28.78 &
& 95.77 & 18.40 &
& 95.55 & 19.36 \\

&{SCALE~\cite{Xu:2024:STT}} 
& 98.17 & 9.50 &
& 95.02 & 23.27 &
& 92.26 & 34.51 &
& 97.37 & 12.93 & 
& 95.71 & 20.05 \\  

&{NN-SCALE} 
& 98.04 & 10.08 &
& 94.72 & 24.12 &
& 91.74 & 35.58 &
& 97.53 & 12.50 & 
& 95.51 & 20.57 \\

\cmidrule(l{6pt}r{0pt}){2-16}

 &{$\accentset{\leftarrow}{S}$} (ours)
& \bfseries98.67 & \bfseries6.71 &
& \bfseries96.48 & \bfseries15.16 &
& \bfseries95.05 & \bfseries21.95 &
& 89.09 & 43.48 & 
& 94.80 & 21.82 \\

 &{$\combi{S}$} {\bfseries ActSub} (ours)
& 98.51 & 7.19 &
& 96.40 & 16.73 &
& 94.40 & 25.64 &
& 97.66 & 11.01 &  
& \bfseries96.74 & \bfseries15.14 \\

    \bottomrule
  \end{tabularx}
  \vspace{-1.0em}
  \caption{\textit{OOD detection accuracies for ResNet-50 trained on ImageNet-1k} (all in \%). We consider three model families. For baseline methods that are mentioned multiple times, we report the version that corresponds to each model family. For each setting, we select the top results from the respective paper (\eg, for fDBD \cite{Liu:2024:FBD}, we consider the variant with SCALE \cite{Xu:2024:STT}). %
  }
  \vspace{-1.5em}
  \label{tab:results_main_resnet50}
\end{table*}

%% file: tables/mobilenet.tex
\begin{table*}
  \centering
  \scriptsize
  \begin{tabularx}{\textwidth}{@{}X SS R SS R SS  R SS R SS@{}}
    \toprule
    \multicolumn{1}{X}{} 
    & \multicolumn{2}{c}{\phantom{xii}\textbf{iNaturalist}} &
    & \multicolumn{2}{c}{\textbf{SUN}} &
    & \multicolumn{2}{c}{\textbf{Places}} &
    & \multicolumn{2}{c}{\textbf{Textures}} &
    & \multicolumn{2}{c}{\textbf{Average}}\\
 
\cmidrule(l{6pt}r{0pt}){2-3} \cmidrule(l{0pt}r{0pt}){5-6} \cmidrule(l{0pt}r{0pt}){8-9} \cmidrule(l{0pt}r{0pt}){11-12} \cmidrule(l{0pt}r{0pt}){14-15}

{\textbf{Method}} 
& {AUC $(\uparrow)$} & {FPR $(\downarrow)$} &
& {AUC $(\uparrow)$} & {FPR $(\downarrow)$} &
& {AUC $(\uparrow)$} & {FPR $(\downarrow)$} &
& {AUC $(\uparrow)$} & {FPR $(\downarrow)$} &
& {AUC $(\uparrow)$} & {FPR $(\downarrow)$} \\

\midrule

{MSP~\cite{Hendrycks:2017:BDM}} 
& 85.71 & 63.09 & 
& 76.01 & 79.67 & 
& 75.51 & 81.47 & 
& 76.49 & 75.12 & 
& 78.43 & 74.84 \\

{Energy~\cite{Liu:2020:EOD}} 
& 88.91 & 59.50 &
 & 84.50 & 62.65 &
& 81.19 & 69.37 &
& 85.03 & 58.05 & 
& 84.91 & 62.39\\

{ODIN~\cite{Liang:2018:ERO}} 
& 91.33 & 45.61 & 
& 83.44 & 63.03 & 
& 80.85 & 70.01 & 
& 85.61 & 52.45 & 
& 85.31 & 57.78 \\

{ReAct~\cite{Sun:2021:ODR}} 
& 92.72 & 43.07 &
& 87.26 & 52.47 &
& 84.07 & 59.91 &
& 90.96 & 40.20 & 
& 88.75 & 48.91\\

{BATS~\cite{Zhu:2022:BOD}} 
& 94.33 & 31.56 &
& 90.21 & 41.68 &
& 86.26 & 52.43 &
& 90.76 & 38.69 & 
& 90.39 & 41.09\\

{DICE~\cite{Sun:2022:LSO}} 
& 93.57 & 32.30  &
 & 92.86 & 31.22  &
& 88.02 & 46.78 &
& 96.25 & 16.28  & 
& 92.68 & 31.64\\
       
{NRE~\cite{Guan:2023:RPT}} 
& 93.66 & 35.84 &
& 90.77 & 40.35 &
& 86.76 & 52.38 &
& 95.39 & 18.44 & 
& 91.65 & 36.75\\

{LINe~\cite{Ahn:2023:ODL}} 
& 95.53 & 24.95 &
& 92.94 & 33.19 &
& 88.98 & 47.95 &
& 97.05 & 12.30 & 
& 93.62 & 29.60\\

{ASH-S~\cite{Djurisic:2023:ESA}} 
&  91.94 & 39.10 &
 & 90.02 & 43.62 &
& 84.73 & 58.84 &
& \bfseries97.10 & \bfseries13.12  & 
& 90.95 & 38.67\\
        
{NNGuide~\cite{Park:2023:NNG}} 
& 91.19 & 45.73 &
& 87.87 & 51.03 &
& 84.44 & 60.60 &
& 92.47 & 29.50 & 
& 88.99 & 46.72\\

{CoRP~\cite{Fang:2024:KPO}} 
& 94.27& 31.72 &
& 90.98 & 40.77 &
& 86.42 & 55.69 &
& 97.49 & 10.48  & 
& 92.29 & 34.66 \\

{DDCS~\cite{Yuan:2024:DDC}} 
& 96.87& 17.44 &
 & 95.83 & 17.42 &
& 91.80 & 30.49 &
& 94.86 & 25.11  & 
& 94.84 & 22.61 \\

{SCALE~\cite{Xu:2024:STT}} 
& 94.43 & 30.26 &
 & 91.67 & 38.56 &
& 87.35 & 53.28 &
& 96.66 & 14.63  & 
& 92.52 & 34.18 \\  
      
\midrule

{$\accentset{\rightarrow}{S}$} (ours)
& 70.43	 & 82.72 &
 & 73.12	 & 83.53 &
& 64.83	 & 91.89 &
& 96.42	 & 17.27 & 
& 76.20 & 68.85 \\

{$\accentset{\leftarrow}{S}$} (ours)
& \bfseries97.67 & \bfseries12.54&
 & \bfseries96.24 & \bfseries17.30 &
& \bfseries93.51 & \bfseries28.66 &
& 92.58 & 36.21  & 
& 95.00 & 23.68 \\

{$\combi{S}$} {\bfseries ActSub} (ours) 
& 97.43& 13.90 &
& 96.02 & 18.31 &
& 93.08& 30.26	&
& 94.95	 & 24.34 & 
& \bfseries95.37 & \bfseries21.70 \\

    \bottomrule
  \end{tabularx}
  \vspace{-0.5em}
  \caption{\textit{OOD detection accuracies for MobileNetV2 trained on ImageNet-1k} (all in \%). We report each of our components separately and in combination. For baseline methods, we consider best-reported results with this setting, \eg, in combination with a previous method. }
  \vspace{-
  1.0em}
  \label{tab:results_main_mobilenet}
\end{table*}

%% file: tables/cifar_main.tex
\begin{table}
  \centering
  \scriptsize
  \begin{tabularx}{\columnwidth}{@{}X SS R SS@{}}
    \toprule
    \multicolumn{1}{X}{} 
    & \multicolumn{2}{c}{\phantom{xii}\textbf{CIFAR10}} &
    & \multicolumn{2}{c}{\textbf{CIFAR100}}\\
 
\cmidrule(l{6pt}r{0pt}){2-3} \cmidrule(l{6pt}r{0pt}){4-6} 

{\textbf{Method}} 
& {AUC $(\uparrow)$} & {FPR $(\downarrow)$} &
& {AUC $(\uparrow)$} & {FPR $(\downarrow)$} \\

\midrule

{Energy~\cite{Liu:2020:EOD}} 
& 92.53 & 35.55 &

& 77.39  & 81.34  \\

{ReAct~\cite{Sun:2021:ODR}} 
& 93.77 & 35.31 &
& 80.24	& 77.38  \\

{DICE~\cite{Sun:2022:LSO}} 
& 93.09 & 30.21 &
& 83.15  & 62.00 \\

{LINe~\cite{Ahn:2023:ODL}} 
& 95.75	& 20.88 &
& 84.63	& 45.74 \\

{ASH-S~\cite{Djurisic:2023:ESA}} 
& 95.25 & 21.12 &
& 87.76  & 47.89 \\

{DDCS~\cite{Yuan:2024:DDC}} 
& 96.05 & 19.35 &
& 86.73  & 43.05 \\ 

{SCALE~\cite{Xu:2024:STT}} 
& 96.16 & 17.84 &
& 88.44  & 45.86 \\

\midrule
{$\combi{S}$} {\bfseries ActSub} (ours) 
& \bfseries96.87 & \bfseries15.37 &

& \bfseries89.35  & \bfseries41.26 \\

    \bottomrule
  \end{tabularx}
  \vspace{-1.0em}
  \caption{\textit{CIFAR OOD detection accuracies for DenseNet-101} (all in \%). We report average accuracy over four OOD datasets when CIFAR10 and CIFAR100 are ID.}
  \label{tab:cifar_main}
  \vspace{-0.5em}
  
\end{table}

%% file: tables/openood_main.tex
\begin{table}
  \centering
  \scriptsize
  \begin{tabularx}{\columnwidth}{@{}X SS R SS@{}}
    \toprule
    \multicolumn{1}{X}{} 
    & \multicolumn{2}{c}{\phantom{xii}\textbf{Near-OOD (Avg.)}} &
    & \multicolumn{2}{c}{\textbf{Far-OOD (Avg.)}}\\
 
\cmidrule(l{6pt}r{0pt}){2-3} \cmidrule(l{6pt}r{0pt}){4-6} 

{\textbf{Method}} 
& {AUC $(\uparrow)$} & {FPR $(\downarrow)$} &
& {AUC $(\uparrow)$} & {FPR $(\downarrow)$} \\

\midrule

{MSP~\cite{Hendrycks:2017:BDM}} 
& 76.02 & 65.67 &
& 85.23  & 51.47  \\
{Energy~\cite{Liu:2020:EOD}} 
& 75.89 & 68.56 &
& 89.47  & 38.40  \\

{ReAct~\cite{Sun:2021:ODR}} 
& 77.38 & 66.75 &
& 93.67  & 26.31 \\

{RankFeat~\cite{Koyejo:2022:RFR}} 
& 50.99 & 91.83 &
& 53.93 & 87.17 \\
 
{ViM~\cite{Sun:2022:ODD}} 
& 72.08  & 71.35 &
& 92.68  & 24.67 \\

{SHE~\cite{Zhang:2023:ODI}} 
&  73.78 & 73.01  &
& 90.92  & 41.45 \\

{GEN~\cite{Liu:2023:PLS}} 
&  76.85 & 65.32  &
& 89.76  & 35.61 \\

{ASH-S~\cite{Djurisic:2023:ESA}} 
&  79.63 & 62.03  &
& 96.47  & 16.86 \\ 

{WeiPer+KLD~\cite{Granz:2024:DWP}} 
&  80.05 & 61.39  &
& 95.54  & 22.08 \\ 
  
{SCALE~\cite{Xu:2024:STT}} 
& 81.36 & 59.76 &
& 96.53  & 16.53 \\ 
\midrule

{$\combi{S}$ {\bfseries ActSub} w/ ReAct}
& 77.70 & 65.36 &
& 96.56  & 15.57 \\
{$\combi{S}$ {\bfseries ActSub} w/ ASH-S}
& 84.06 & \bfseries52.46 &
& 96.91 & \bfseries14.22 \\
{$\combi{S}$ {\bfseries ActSub} w/ SCALE}
& \bfseries84.24 & 52.60 &
& \bfseries96.96  & 14.29 \\

    \bottomrule
  \end{tabularx}
  \vspace{-1.0em}
  \caption{\textit{OpenOOD OOD detection accuracies for ResNet-50 trained on ImageNet-1k} (all in \%). We report average Near- and Far-OOD accuracy. %
  }
  \vspace{-1.0em}
  \label{tab:results_openood}
\end{table}

%% file: tables/pca_ablation.tex
\begin{table}
  \centering
  \scriptsize
  \begin{tabularx}{\columnwidth}{@{}cX SS R SS@{}}
    \toprule
    &\multicolumn{1}{X}{} 
    & \multicolumn{2}{c}{\phantom{xii}\textbf{Near-OOD (Avg.)}} &
    & \multicolumn{2}{c}{\textbf{Far-OOD (Avg.)}}\\
 
\cmidrule(l{6pt}r{0pt}){3-4} \cmidrule(l{6pt}r{0pt}){5-7} 

&{\textbf{Basis}} 
& {AUC $(\uparrow)$} & {FPR $(\downarrow)$} &
& {AUC $(\uparrow)$} & {FPR $(\downarrow)$} \\

\midrule
\parbox[t]{1mm}{\multirow{4}{*}{\rotatebox[origin=c]{0}{$\accentset{\rightarrow}{S}$ }}}

&{PCA} 
& 64.10	& \bfseries75.84 &
& \cellcolor{gray!20}71.55 & \multicolumn{1}{>{\columncolor{gray!20}}c}{63.00}  \\

&{SI-PCA} 
& 65.68 & 78.42 &
& \cellcolor{gray!20}90.62  & \multicolumn{1}{>{\columncolor{gray!20}}c}{34.94}  \\

&{Null Space} 
& 66.43	 & 77.64 &
& \cellcolor{gray!20}92.20  & \multicolumn{1}{>{\columncolor{gray!20}}c}{31.11} \\

&{SVD} 
&  \bfseries67.51 & 76.27  &
&  \cellcolor{gray!20}\bfseries92.74 & \multicolumn{1}{>{\columncolor{gray!20}}c}{\bfseries29.51}  \\

\midrule
\parbox[t]{1mm}{\multirow{4}{*}{\rotatebox[origin=c]{0}{$\accentset{\leftarrow}{S}$ }}}

&{PCA} 
& \cellcolor{gray!20}80.97	 & \cellcolor{gray!20}60.30 &
& 96.10  & \multicolumn{1}{>{\columncolor{gray!20}}c}{18.75}  \\

&{SI-PCA} 
& \cellcolor{gray!20}81.37 & \cellcolor{gray!20}60.06 &
& \bfseries96.36  & \multicolumn{1}{>{\columncolor{gray!20}}c}{\bfseries17.54}  \\

&{Null Space} 
& \cellcolor{gray!20}81.68	 & \cellcolor{gray!20}59.71 &
& 96.35	 & \multicolumn{1}{>{\columncolor{gray!20}}c}{17.80} \\
 
&{SVD} 
& \cellcolor{gray!20}\bfseries84.90 & \cellcolor{gray!20}\bfseries56.39	 &
& \cellcolor{red!20}93.58  & \multicolumn{1}{>{\columncolor{red!20}}c}{29.31} \\

\midrule
\parbox[t]{1mm}{\multirow{4}{*}{\rotatebox[origin=c]{0}{$\combi{S}$ }}}

&{PCA} 
& \cellcolor{gray!20}80.40	 & \cellcolor{gray!20}58.44 &
& \cellcolor{gray!20}96.17  & \multicolumn{1}{>{\columncolor{gray!20}}c}{17.30}  \\

&{SI-PCA} 
& \cellcolor{gray!20}80.69	 & \cellcolor{gray!20}59.99 &
& \cellcolor{gray!20}96.73	 & \multicolumn{1}{>{\columncolor{gray!20}}c}{15.43}  \\

&{Null Space} 
& \cellcolor{gray!20}80.04	 & \cellcolor{gray!20}60.94 &
& \cellcolor{gray!20}96.78  & \multicolumn{1}{>{\columncolor{gray!20}}c}{15.10} \\

&{SVD} 
& \cellcolor{gray!20}\bfseries84.24 & \cellcolor{gray!20}\bfseries52.60  &
& \cellcolor{gray!20}\bfseries96.96  & \multicolumn{1}{>{\columncolor{gray!20}}c}{\bfseries14.29}\\

    \bottomrule
  \end{tabularx}
  \vspace{-1.0em}
  \caption{
  \textit{Different bases for} $\protect\accentset{\rightarrow}{S}$, $\protect\accentset{\leftarrow}{S}$, and $\protect\combi{S}$.
  Our proposed SVD basis outperforms other bases for $\protect\accentset{\rightarrow}{S}$ on the Far-OOD setting and for $\protect\accentset{\leftarrow}{S}$ in the Near-OOD setting \emph{(gray)}. However, $\protect\accentset{\leftarrow}{S}$ with SVD suffers in the Far-OOD setting \emph{(red)}. When combining both score functions, SVD leads in both settings.
  }
  \vspace{-1.0em}
  \label{tab:pca_ablation_openood}
\end{table}

%% file: tables/component_ablation.tex
\begin{table}
  \centering
  \scriptsize
  \begin{tabularx}{\columnwidth}{@{}cX SS T SS@{}}
    \toprule
    &\multicolumn{1}{X}{} 
    & \multicolumn{2}{c}{\phantom{xii}\textbf{Near-OOD (Avg.)}} &
    & \multicolumn{2}{c}{\textbf{Far-OOD (Avg.)}}\\
 
\cmidrule(l{6pt}r{0pt}){3-4} \cmidrule(l{6pt}r{0pt}){5-7} 

&{\textbf{Component for $\accentset{\rightarrow}{S}$}} 
& {AUC $(\uparrow)$} & {FPR $(\downarrow)$} &
& {AUC $(\uparrow)$} & {FPR $(\downarrow)$} \\

\midrule

\parbox[t]{1mm}{\multirow{3}{*}{\rotatebox[origin=c]{0}{$\accentset{\rightarrow}{S}$}}}

&{$\decisive{\mathbf{a}}$} 
& 70.18 & 73.74 &
&76.34 & 60.50
  \\
		
&{${\mathbf{a}}$} 
&   \bfseries71.75&  \bfseries70.11 &
& 87.77	 &  40.80 \\
		
&{$\insignificant{\mathbf{a}}$} 
& 67.51	& 76.27 &
&  \bfseries92.74	&  \bfseries29.30 \\

\midrule
\parbox[t]{1mm}{\multirow{3}{*}{\rotatebox[origin=c]{0}{$\combi{S}$}}}

&{$\decisive{\mathbf{a}}$} 
& 81.89	& 56.77	 &
& 92.47	 & 28.41 \\

&{${\mathbf{a}}$} 
& 84.11 & 52.86	 &
& 96.00	& 17.98 \\

&{$\insignificant{\mathbf{a}}$} 
&  \bfseries84.24 & \bfseries52.60  &
& \bfseries96.96  & \bfseries14.29 \\

    \bottomrule
  \end{tabularx}
  \vspace{-1.0em}
  \caption{\textit{Different components for} $\protect\accentset{\rightarrow}{S}$.
  We show how different activation components ($\protect\decisive{\mathbf{a}}$, $\protect{\mathbf{a}}$, $\protect\insignificant{\mathbf{a}}$) affect our cosine similarity-based score function $\protect\accentset{\rightarrow}{S}$, when considered in isolation and in combination with the activation shaping-based $\protect\accentset{\leftarrow}{S}$ (\ie, $\protect\combi{S}$).%
  }
  \vspace{-0.5em}
  \label{tab:part_ablation}
\end{table}

%% file: sec/5_conclusion.tex
\section{Conclusion}
\label{sec:conclusion}

OOD detection methods rely on the %
information contained in the activation space. %
In this study, we define two orthogonal subspaces within the activation space: \emph{decisive} and \emph{insignificant}. The decisive component of activations determines the model's decision. On the other hand, the insignificant component is decoupled from the classification output and yields powerful features, akin to a random neural network, that are discriminative for OOD detection. 
Further, we observe that selectively applying activation shaping to the decisive component increases the separability of ID and OOD distributions, particularly for OOD inputs close to the ID domain.
By combining discriminative information from both subspaces, our method ActSub outperforms relevant baselines across a variety of different benchmarks.%

\myparagraph{Acknowledgments.} This project has received funding from the European Research Council (ERC) under the European
Union’s Horizon 2020 research and innovation programme (grant agreement No.\ 866008).
SR further acknowledges support by the Deutsche Forschungsgemeinschaft (German Research Foundation, DFG) under Germany’s Excellence Strategy (EXC 3066/1 “The Adaptive Mind”, Project No.\ 533717223).

%% file: sec/X_suppl_cr.tex
\clearpage
\setcounter{section}{0}
\renewcommand\thesection{\Alph{section}}
\setcounter{page}{1}
\pagenumbering{roman}
\maketitlesupplementary

\begin{figure}[h]
\centering
\input{supp/figures/dist_with_gaussian}
\vspace{-0.5em}
\caption{\textit{Score distributions of ID and OOD data.} We fit multivariate Gaussians to the concatenated (along a new second dimension) score values from the decisive ($\protect\decisive{S}$) and insignificant ($\protect\insignificant{S}$) components of the sampled ID (ImageNet-1k \cite{Deng:2009:LHI, Russakovsky:2015:INS}), Near-OOD (NINCO~\cite{Bitterwolf:2023:FIO}), and Far-OOD (Textures~\cite{Cimpoi:2014:DTW}) data. We use ResNet-50 \cite{He:2016:DRL}. Score values are negated to visualize in the positive domain.}
\label{fig:const}
\vspace{-0.5em}
\end{figure}
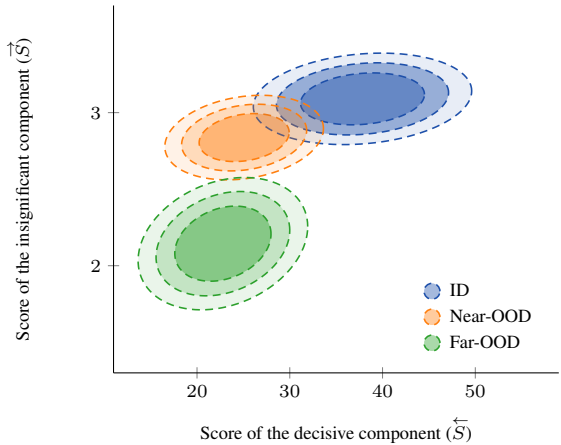

\section{Additional Experiments}\label{sec:add_ablt}
We extend our experiments in the setting of \cref{subsec:ablation}. We use the OpenOOD~\cite{Zhang:2023:EBO} benchmark and report Near-OOD and Far-OOD results separately.

\myparagraph{ViT.} \cref{tab:rebuttal_vit} reports the accuracy of our method ActSub with ViT-B/16 \cite{Dosovitskiy:2021:ICLR}, ResNet-50, and their average. Since activation shaping methods (including SCALE) are known to perform worse on ViTs \cite{Sun:2021:ODR, Djurisic:2023:ESA, Xu:2024:STT}, we additionally combine ActSub with GEN \cite{Liu:2023:PLS} on the decisive component. For ViT, we use the prototype variant of ID data (see below: ``Dependence on the data volume'') for the insignificant component, which we empirically found to work better. For ViT, ActSub improves GEN and SCALE \cite{Xu:2024:STT}, particularly in Near-OOD, improving the AUC by \SI{2.15}{\percent}/\SI{18.09}{\percent} and reducing the FPR by \SI{6.68}{\percent}/\SI{28.10}{\percent}, respectively. Remarkably, when considering the average of ResNet and ViT, our variant with SCALE reaches SotA accuracy. %

\myparagraph{Visualization of the score distributions.} \cref{fig:const} visualizes the score distributions of our decisive ($\decisive{S}$) and insignificant ($\insignificant{S}$) components for ID, Near-OOD, and Far-OOD data. For ID, Near-OOD, and Far-OOD, we use ImageNet-1k~\cite{Deng:2009:LHI, Russakovsky:2015:INS}, NINCO~\cite{Bitterwolf:2023:FIO}, and Textures~\cite{Cimpoi:2014:DTW}, respectively. We observe that the Far-OOD data is better separated from the ID data with the score of the insignificant component. Inversely, Near-OOD and ID are better separated by the score of the decisive component. With this observation, we emphasize the complementary effect of our two score functions.

\begin{figure}[t]
\centering
\input{supp/figures/k_abl}
\vspace{-1.5em}
\caption{\textit{Classification and OOD detection accuracy (\%, $\uparrow$) for different ratios of the norms of decisive and insignificant components with ResNet (orange) and MobileNet (blue).} The AUC is calculated by the average of datasets in OpenOOD \cite{Zhang:2023:EBO}, and ID is ImageNet-1k \cite{Deng:2009:LHI, Russakovsky:2015:INS}. }
\label{fig:k_up}
\vspace{-0.5em}
\end{figure}
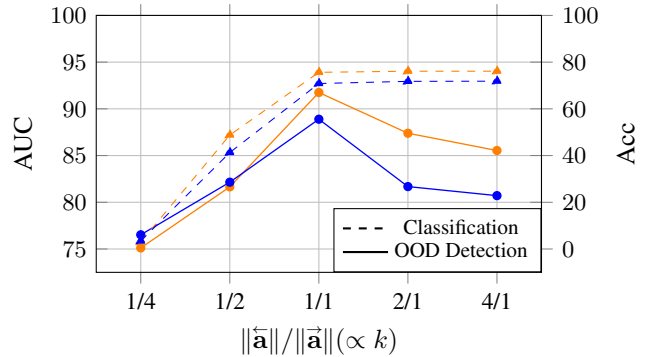

\myparagraph{Choice of hyperparameter $k$.}
Ideally, the decisive subspace exclusively captures the classification signal while the insignificant subspace captures all information that does not aid classification. We plot the OOD-detection and ID-classification accuracy against the norm ratio of the two components -- which is proportional to $k$ -- in \cref{fig:k_up}. For ID, we use ImageNet-1k \cite{Deng:2009:LHI, Russakovsky:2015:INS}), and for OOD, we consider the average accuracy of the datasets in OpenOOD~\cite{Zhang:2023:EBO}. The ID-classification accuracy already saturates with a balanced decomposition where the norms of the two components are equal (1/1), corresponding to the ratio we propose in our experiments. Shifting the balance towards $\insignificant{\mathbf{a}}$ (smaller $k$), causes classification-related directions to be captured in the insignificant subspace, which has a detrimental effect on OOD detection (\cf \cref{tab:part_ablation}). Inversely, shifting the balance towards $\decisive{\mathbf{a}}$ (larger $k$) results in insignificant directions being captured by the decisive subspace, and consequently, increasing interference. To summarize, deviations in either direction from the 1/1 ratio reduce the OOD-detection accuracy, confirming our choice in \cref{eq:ratio}. 

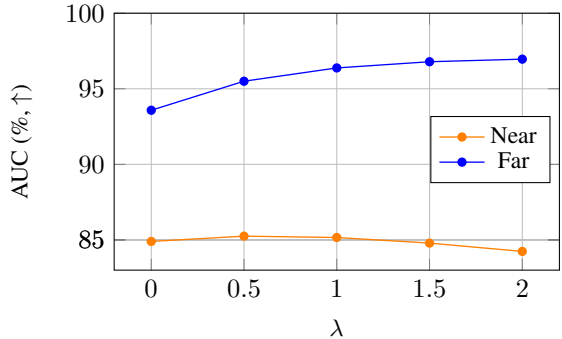
\begin{figure}[t]
\centering
\input{supp/figures/alpha_ablation}
\vspace{-0.5em}
\caption{\textit{Sensitivity of hyperparameter $\lambda$.} We report the AUC for our method ActSub ($\protect\combi{S}$) for different $\lambda$ values for Near-OOD and Far-OOD of the OpenOOD~\cite{Zhang:2023:EBO} benchmark. We use a ResNet-50~\cite{He:2016:DRL} model trained on ImageNet-1k~\cite{Deng:2009:LHI, Russakovsky:2015:INS}.}
\label{fig:alpha_abl}
\vspace{-0.5em}
\end{figure}

\myparagraph{Sensitivity of hyperparameter $\lambda$.}
The hyperparameter $\lambda$ weights the information of the decisive component ($\decisive{S}$) that excels in Near-OOD and the information of the insignificant component ($\insignificant{S}$) that excels in Far-OOD. In other words, $\lambda$ balances the discriminative information of each subspace that targets a different aspect of the distribution shift. 
In practice, not only the scale but also the variation and the margin of each score distribution might vary. To account for this, we define $\lambda$ as an exponent (instead of a simple weighting factor) and ensure that the score function of each component contributes sufficiently to the final score ($\combi{S}$). 

We investigate the effect of the hyperparameter $\lambda$ on the accuracy of our method ActSub in \cref{fig:alpha_abl}. Note that $\lambda=0$ is equivalent to only using the score of the decisive component. We observe that Near-OOD performance is relatively unaffected by $\lambda$. On the other hand, for Far-OOD, introducing the cosine similarity-based score of our insignificant component with increasing $\lambda$ significantly increases the accuracy.
We use $\lambda = 0.5, 1, 2$ for MobileNetV2~\cite{Sandler:2018:IRL}, ViT, and ResNet-50, respectively. For ViT with GEN \cite{Liu:2023:PLS}, we use $\lambda = 0.5$. In the main experiments with ImageNet-1k~\cite{Deng:2009:LHI, Russakovsky:2015:INS} (\cf \cref{tab:results_main_resnet50,tab:results_main_mobilenet}), we use NINCO~\cite{Bitterwolf:2023:FIO} as the held-out validation set. For CIFAR~\cite{Krizhevsky:2009:LML}, we use MNIST~\cite{MNIST}. For OpenOOD~\cite{Zhang:2023:EBO} experiments, we use the OpenOOD validation split.

\myparagraph{Sensitivity of the decisive component.}
We design the decisive component ($\decisive{S}$) to capture the information related to the classification task, which we assume to be mostly semantic information. To reflect this, when tuning the hyperparameters of the decisive component, \ie, the hyperparameters of SCALE \cite{Xu:2024:STT} or GEN \cite{Liu:2023:PLS}, we incorporate ImageNet-R \cite{Hendrycks:2021:MFR}, which includes stylized renditions of ImageNet classes, along with ImageNet-1k as an ID dataset. By doing so, we introduce non-semantic variations to the ID data and guide the decisive component to focus on the semantic variations between ID and OOD distributions. We select the parameter that maximizes the difference between AUC and FPR.

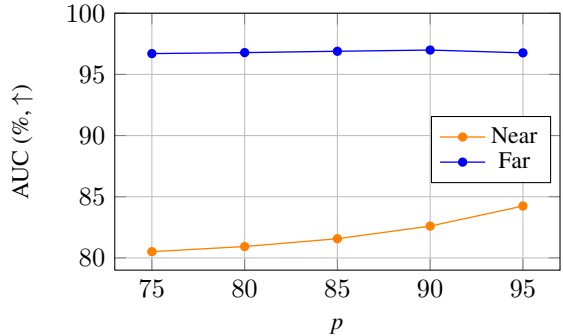
\begin{figure}[t]
\centering
\input{supp/figures/prune_ablation}
\vspace{-0.5em}
\caption{\textit{Sensitivity of pruning percentage $\textit{p}$ of SCALE.} We report the AUC for our method ActSub ($\protect\combi{S}$) for different percentages of pruning ($\textit{p}$) for Near-OOD and Far-OOD from the OpenOOD~\cite{Zhang:2023:EBO} benchmark. We use a ResNet-50~\cite{He:2016:DRL} model trained on ImageNet-1k~\cite{Deng:2009:LHI, Russakovsky:2015:INS}.}
\label{fig:p_abl}
\vspace{-0.5em}
\end{figure}

In \cref{fig:p_abl}, we present how the pruning percentage $\textit{p}$ of the activation shaping method SCALE \cite{Xu:2024:STT} affects the accuracy of our combined score function ($\combi{S}$). For Far-OOD, the effect of $\textit{p}$ is insignificant. However, for Near-OOD, the accuracy increases with increasing $\textit{p}$. For the original SCALE, \ie, when applied on the whole activation, the accuracy has been shown to decrease after \SI{85}{\percent} \cite{Xu:2024:STT}. We believe this is because SCALE needs to keep more channels to capture discriminative information when applied to the whole representation. In contrast, when only applied to the decisive component ($\decisive{S}$), SCALE benefits from higher pruning percentages -- with the interference of the insignificant component being removed from the activation, a higher pruning percentage still captures the important channels for the model's prediction.

\input{tables/data_main}

\myparagraph{Dependence on the data volume.}
\cref{tab:sup_data} shows the OOD detection accuracy of our method using different amounts of ID data used to compute the cosine similarity of the insignificant component. Compared to 10\% used in our main method, reducing the data volume to 1\% either by random sampling or sample averaging maintains strong accuracy. Furthermore, we evaluate a prototype approach, where each sample is a cluster center obtained by applying $k$-means to the insignificant component sampled from the training split. The number of prototypes corresponds to 0.1\% of the training data volume. This shows that we can \emph{(i)} significantly reduce memory requirements with a sparse set of representations (0.1\% of the data volume), \emph{(ii)} reduce privacy vulnerability, as reconstructing individual samples from cluster centers is harder, and \emph{(iii)} retain the significant improvement over previous methods (\cf \cref{tab:results_openood}).

\input{tables/vit_full_main}

\section{Expanded Tables}\label{sec:exp_tab}
We provide an expanded version of the tables (\cref{tab:results_openood,tab:cifar_main}) for our $\text{CIFAR}$~\cite{Krizhevsky:2009:LML} and OpenOOD~\cite{Zhang:2023:EBO} experiments in \cref{subsec:further_exp}. \cref{tab:cifar_exp} shows the accuracy of our method ActSub compared to baselines for each separate OOD dataset for when CIFAR10 and CIFAR100 are ID. Similarly, in \cref{tab:openood_exp}, we present the accuracy of ActSub for each individual dataset in Near-OOD and Far-OOD settings of the OpenOOD benchmark.

\input{supp/tables/cifar_expand}

\input{supp/tables/openood_expand}

%% file: supp/figures/dist_with_gaussian.tex
\definecolor{IDcolor}{RGB}{30, 80, 170}
\definecolor{NearOOD}{RGB}{255, 127, 14}
\definecolor{FarOOD}{RGB}{44, 160, 44}

\begin{tikzpicture}
    \scriptsize
    \begin{axis}[
        axis lines=left,                    %
        xlabel={Score of the decisive component (${\decisive{S}}$}), ylabel={Score of the insignificant component (${\insignificant{S}}$}),         %
        xmin=15, xmax=55, ymin=1.5, ymax=3.5,    %
        samples=100, samples y=100,          %
        enlarge x limits, enlarge y limits,   %
        axis line style=solid, 
        width=0.9\linewidth,
        axis line style={-}, %
        legend pos=south east,
        legend cell align={left},
        legend style={draw=none, font=\scriptsize}, 
        legend image post style={only marks, mark=*, mark size=3pt}
    ]   
        \def\rotationAngle{0.25}
        \def\rotationAngleT{0.4}
        \def\rotationAngleN{0.7}%
        
        \def\sigmaX{10.00} %
        \def\sigmaY{0.25} %
        
        \def\muX{37.84} %
        \def\muY{3.09}
        
        \def\sigmaXT{7.263721655681976} %
        \def\sigmaYT{0.2290828362678113} %
        
        \def\muXT{25.08091} %
        \def\muYT{2.837134}

        \def\sigmaXN{7.765667897851857} %
        \def\sigmaYN{0.35472169121774666} %
        
        \def\muXN{22.790535} %
        \def\muYN{2.1440575}

        \addplot[IDcolor, semithick, fill=IDcolor, fill opacity=0.4, densely dashed, domain=0:360] 
            ({\muX + \sigmaX * cos(x) * sqrt(-2 * ln(0.8)) * cos(\rotationAngle) - \sigmaY * sin(x) * sqrt(-2 * ln(0.8)) * sin(\rotationAngle)}, 
             {\muY + \sigmaX * cos(x) * sqrt(-2 * ln(0.8)) * sin(\rotationAngle) + \sigmaY * sin(x) * sqrt(-2 * ln(0.8)) * cos(\rotationAngle)});

        \addplot[NearOOD, semithick, fill=NearOOD, fill opacity=0.4, densely dashed, domain=0:360] 
            ({\muXT + \sigmaXT * cos(x) * sqrt(-2 * ln(0.8)) * cos(\rotationAngleT) - \sigmaYT * sin(x) * sqrt(-2 * ln(0.8)) * sin(\rotationAngleT)}, 
             {\muYT + \sigmaXT * cos(x) * sqrt(-2 * ln(0.8)) * sin(\rotationAngleT) + \sigmaYT * sin(x) * sqrt(-2 * ln(0.8)) * cos(\rotationAngleT)});

        \addplot[FarOOD, semithick, fill=FarOOD, fill opacity=0.4, densely dashed, domain=0:360] 
            ({\muXN + \sigmaXN * cos(x) * sqrt(-2 * ln(0.8)) * cos(\rotationAngleN) - \sigmaYN * sin(x) * sqrt(-2 * ln(0.8)) * sin(\rotationAngleN)}, 
             {\muYN + \sigmaXN * cos(x) * sqrt(-2 * ln(0.8)) * sin(\rotationAngleN) + \sigmaYN * sin(x) * sqrt(-2 * ln(0.8)) * cos(\rotationAngleN)});

        \addplot[IDcolor, semithick, fill=IDcolor, fill opacity=0.1, densely dashed, domain=0:360] 
            ({\muX + \sigmaX * cos(x) * sqrt(-2 * ln(0.5)) * cos(\rotationAngle) - \sigmaY * sin(x) * sqrt(-2 * ln(0.5)) * sin(\rotationAngle)}, 
             {\muY + \sigmaX * cos(x) * sqrt(-2 * ln(0.5)) * sin(\rotationAngle) + \sigmaY * sin(x) * sqrt(-2 * ln(0.5)) * cos(\rotationAngle)});
        \addplot[IDcolor, semithick, fill=IDcolor, fill opacity=0.4, densely dashed, domain=0:360] 
            ({\muX + \sigmaX * cos(x) * sqrt(-2 * ln(0.65)) * cos(\rotationAngle) - \sigmaY * sin(x) * sqrt(-2 * ln(0.65)) * sin(\rotationAngle)}, 
             {\muY + \sigmaX * cos(x) * sqrt(-2 * ln(0.65)) * sin(\rotationAngle) + \sigmaY * sin(x) * sqrt(-2 * ln(0.65)) * cos(\rotationAngle)});

        \addplot[NearOOD, semithick, fill=NearOOD, fill opacity=0.1, densely dashed, domain=0:360] 
            ({\muXT + \sigmaXT * cos(x) * sqrt(-2 * ln(0.5)) * cos(\rotationAngleT) - \sigmaYT * sin(x) * sqrt(-2 * ln(0.5)) * sin(\rotationAngleT)}, 
             {\muYT + \sigmaXT * cos(x) * sqrt(-2 * ln(0.5)) * sin(\rotationAngleT) + \sigmaYT * sin(x) * sqrt(-2 * ln(0.5)) * cos(\rotationAngleT)});

        \addplot[NearOOD, semithick, fill=NearOOD, fill opacity=0.2, densely dashed, domain=0:360] 
            ({\muXT + \sigmaXT * cos(x) * sqrt(-2 * ln(0.65)) * cos(\rotationAngleT) - \sigmaYT * sin(x) * sqrt(-2 * ln(0.65)) * sin(\rotationAngleT)}, 
             {\muYT + \sigmaXT * cos(x) * sqrt(-2 * ln(0.65)) * sin(\rotationAngleT) + \sigmaYT * sin(x) * sqrt(-2 * ln(0.65)) * cos(\rotationAngleT)});

        \addplot[FarOOD, semithick, fill=FarOOD, fill opacity=0.1, densely dashed, domain=0:360] 
            ({\muXN + \sigmaXN * cos(x) * sqrt(-2 * ln(0.5)) * cos(\rotationAngleN) - \sigmaYN * sin(x) * sqrt(-2 * ln(0.5)) * sin(\rotationAngleN)}, 
            {\muYN + \sigmaXN * cos(x) * sqrt(-2 * ln(0.5)) * sin(\rotationAngleN) + \sigmaYN * sin(x) * sqrt(-2 * ln(0.5)) * cos(\rotationAngleN)});

        \addplot[FarOOD, semithick, fill=FarOOD, fill opacity=0.2, densely dashed, domain=0:360] 
            ({\muXN + \sigmaXN * cos(x) * sqrt(-2 * ln(0.65)) * cos(\rotationAngleN) - \sigmaYN * sin(x) * sqrt(-2 * ln(0.65)) * sin(\rotationAngleN)}, 
            {\muYN + \sigmaXN * cos(x) * sqrt(-2 * ln(0.65)) * sin(\rotationAngleN) + \sigmaYN * sin(x) * sqrt(-2 * ln(0.65)) * cos(\rotationAngleN)});

    \legend{
        {\color{IDcolor}} ID, 
        {\color{NearOOD}} Near-OOD, 
        {\color{FarOOD}} Far-OOD 
    }
        
    \end{axis}

\end{tikzpicture}

%% file: supp/figures/k_abl.tex
\begin{tikzpicture}
        \begin{axis}[
        width=0.9\linewidth,
        height=5.00cm,
        ymin=72.5, ymax=100,
        xmin=-0.5, xmax=4.5,
        xtick={0,1,2,3,4},
        xticklabels={1/4, 1/2, 1/1, 2/1, 4/1},
        ytick={75, 80, 85, 90, 95, 100},
        xlabel={$\| \decisive{\mathbf{a}}\| / \| \insignificant{\mathbf{a}}\| (\propto k)$},
        ylabel={AUC},
        xlabel style={font=\small},
        ylabel style={font=\small},
        legend style={at={(1.0, 0.25)}, anchor=north east, font=\footnotesize, scale=0.3, row sep=-2pt},
        grid=major,
        clip=true,
        every axis plot/.append style={line width=0.5pt, mark=*, mark size=1.5pt},
        xtick align=outside,
        xticklabel style={font=\small},
        yticklabel style={font=\small},
        xlabel near ticks,
        ylabel near ticks,
    ]

    \addplot[black, thick, dashed, mark=none] coordinates {(0,0)}; \addlegendentry{Classification}
    \addplot[black, thick, solid, mark=none] coordinates {(0,0)}; \addlegendentry{OOD Detection}

    \addplot[orange, mark=*] table[x expr=\coordindex, y index=1] {supp/figures/resnet_auc.txt};
    \addplot[blue, mark=*] table[x expr=\coordindex, y index=1] {supp/figures/mobilenet_auc.txt};

    \end{axis}

    \begin{axis}[
        width=0.9\linewidth,
        height=5.00cm,
        ymin=-10, ymax=100,
        xmin=-0.5, xmax=4.5,
        axis y line=right,
        axis x line=none,
        axis line style={-},
        ytick={0, 20, 40, 60, 80, 100},
        ylabel={Acc},
        ylabel style={font\small, rotate=180},
        yticklabel style={font=\small},
        tick align=outside,
        yticklabel style={font=\small},
        legend style={at={(0.98,0.3)}, anchor=north east, font=\tiny, scale=0.3, row sep=-2pt},
        clip=false,
        xlabel near ticks,
        ylabel near ticks,
    ]

    \addplot[orange, dashed, mark=triangle*] table[x expr=\coordindex, y index=1] {supp/figures/resnet_acc.txt};
    \addplot[blue, dashed, mark=triangle*] table[x expr=\coordindex, y index=1] {supp/figures/mobilenet_acc.txt};

    \end{axis}

\end{tikzpicture}

%% file: supp/figures/alpha_ablation.tex
\begin{tikzpicture}
    \begin{axis}[
        width=0.9\linewidth,
        height=5cm,
        xmin=-0.2, xmax=2.2,
        ymin=83, ymax=100,
        xtick={0, 0.5, 1, 1.5, 2},
        ytick={85, 85, 90, 95, 100},
        xlabel={$\lambda$},
        ylabel={AUC (\%, $\uparrow$)},
        xlabel style={font=\small},
        ylabel style={font=\small},
        legend style={at={(0.98,0.6)}, anchor=north east, font=\small, scale=0.5, row sep=-2pt},
        grid=major,
        every axis plot/.append style={line width=0.5pt, mark=*, mark size=1.5pt}
    ]

    \addplot[orange, mark=*] table[x index=0, y index=1] {supp/figures/near_alpha.txt};
    
    \addplot[blue , mark=*] table[x index=0, y index=1] {supp/figures/far_alpha.txt};
    
    \addlegendentry{Near}
    \addlegendentry{Far}

    \end{axis}
\end{tikzpicture}

%% file: tables/data_main.tex
\begin{table}[h!]
  \centering
  \scriptsize
  \renewcommand{\arraystretch}{0.9}
  \begin{tabularx}{\columnwidth}{@{}XXSS SS@{}}
    \toprule
    &
    & \multicolumn{2}{c}{\textbf{Near-OOD (Avg.)}}
    & \multicolumn{2}{c}{\textbf{Far-OOD (Avg.)}}\\
 
\cmidrule(l{6pt}r{0pt}){3-4} \cmidrule(l{6pt}r{0pt}){5-6} 

{\textbf{Volume}} & {\textbf{Sampling}}
& {AUC $(\uparrow)$} & {FPR $(\downarrow)$}
& {AUC $(\uparrow)$} & {FPR $(\downarrow)$} \\

\midrule

{10\%} (\cref{tab:results_openood}) & {Random}
& 84.24 & 52.60 
& 96.96  & 14.29 \\ 
\midrule

{1\%} & {Random}
&  83.33 & 54.99 
& 96.69  & 15.62 \\ 

{1\%} & {Averaging}
&  83.48 & 54.75 
& 96.90  & 14.50 \\ 
  
{0.1\%} & {Clustering}
& 83.73 & 54.41
& 96.85  & 14.77 \\

    \bottomrule
  \end{tabularx}
  \vspace{-0.5em}
  \caption{\textit{Accuracy (in \%) of our unified score function with different data volumes and sampling strategies.} We use ResNet-50 \cite{He:2016:DRL} trained on ImageNet-1k \cite{Deng:2009:LHI, Russakovsky:2015:INS}, and evaluate on OpenOOD \cite{Zhang:2023:EBO}.
  }
  \label{tab:sup_data}
\end{table}

%% file: tables/vit_full_main.tex
\begin{table*}[tbp]
  \centering
  \scriptsize
  \begin{tabularx}{\textwidth}{@{}X R SS R SS R SS@{}}
    \toprule
    \multicolumn{1}{X}{} 

    & & \multicolumn{2}{c}{\phantom{xii}\textbf{ViT-B/16}} && 
     \multicolumn{2}{c}{\textbf{ResNet-50}} & 
    & \multicolumn{2}{c}{\textbf{Avg.\ of ResNet \& ViT}}\\
 
\cmidrule(l{6pt}r{0pt}){3-4} \cmidrule(l{6pt}r{0pt}){6-7}  \cmidrule(l{6pt}r{0pt}){8-10}

{\textbf{Method}}& 
& {Near} & {Far} & 
& {Near} & {Far} & 
& {Near} & {Far}  \\

\midrule

 {RMDS~\cite{Ren:2021:ASF}}& 
& {\bfseries80.09 } / { 65.36} & {92.60 } / { \bfseries28.76} &   
& {76.99 } / { 65.04} & {86.38 } / { 40.91}& 
& {78.54 } / { 65.20} & {89.49 } / { 34.83} \\

 {ViM~\cite{Sun:2022:ODD}}& 
& {77.03 } / { 73.73} & {\bfseries92.84 } / { 29.18}   & 
& {72.08 } / { 71.35} & {92.68 } / { 24.67}& 
& {74.55 } / { 72.54} & {92.76 } / { 26.93} \\ 

 {KNN~\cite{Wang:2022:ODV}}& 
& {74.11 } / { 70.47} & {90.81 } / { 31.93}& 
& {71.10 } / { 70.87} & {90.18 } / { 34.13} & 
& {72.60 } / { 70.67} & {90.50 } / { 33.03} \\ 

 {SHE~\cite{Liu:2023:PLS}}& 
& {76.11 } / { 70.88} & {92.42 } / { 27.12}& 
& {73.78 } / { 73.01} & {90.92 } / { 41.45} & 
& {74.95 } / { 71.94} & {91.67 } / { 34.28} \\ 

 {GEN~\cite{Zhang:2023:ODI}}& 
& {76.30 } / { 70.78} & {91.35 } / { 32.23} & 
& {76.85 } / { 65.32} & {89.79 } / { 35.61}& 
& {76.57 } / { 68.05} & {90.57 } / { 33.92} \\ 
  
 {WeiPer~\cite{Granz:2024:DWP}}& 
& {75.00 } / { 73.02} & {90.32 } / { 38.16} & 
& {80.05 } / { 61.39} & {95.54 } / { 22.08} & 
& {77.52 } / { 67.20} & {92.93 } / { 30.12} \\ 

 {SCALE~\cite{Xu:2024:STT}}& 
& {59.03 } / { 93.94} & {75.22 } / { 86.93}& 
& {81.36 } / { 59.76} & {96.53 } / { 16.53} & 
& {70.19 } / { 76.85} & {85.87 } / { 51.73} \\

\midrule
 {$\combi{S}$ {\bfseries ActSub} w/ GEN}& 
& {78.45 } / { \bfseries64.10} & {91.62 } / { 29.19}& 
& {77.81 } / { 62.97} & {95.90 } / { 18.26} & 
& {78.13 } / { 63.54} & {93.76 } / { 23.73}\\

 {$\combi{S}$ {\bfseries ActSub} w/ SCALE}& 
& {77.12 } / { 65.84} & {90.64 } / { 31.00}  & 
& {\bfseries84.24 } / { \bfseries52.60} & {\bfseries96.96 } / { \bfseries14.29} & 
& {\bfseries80.68 } / { \bfseries59.22} & {\bfseries93.80 } / { \bfseries22.65} \\

    \bottomrule
  \end{tabularx}
  \vspace{-0.5em}
  \caption{\textit{Accuracy of our unified score function reported with ViT and ResNet.} The format is {AUC $(\%, \protect\uparrow)$} / {FPR $(\%, \protect\downarrow)$. Models trained on ImageNet-1k \cite{Deng:2009:LHI, Russakovsky:2015:INS}, and evaluated on OpenOOD \cite{Zhang:2023:EBO}.}
  }
  \label{tab:rebuttal_vit}
\end{table*}

%% file: supp/figures/prune_ablation.tex
\begin{tikzpicture}
    \begin{axis}[
        width=0.9\linewidth,
        height=5cm,
        xmin=73, xmax=97,
        ymin=79, ymax=100,
        xtick={75, 80, 85, 90, 95},
        ytick={80, 85, 90, 95, 100},
        xlabel={\textit{p}},
        ylabel={AUC (\%, $\uparrow$)},
        xlabel style={font=\small},
        ylabel style={font=\small},
        legend style={at={(0.98,0.6)}, anchor=north east, font=\small, scale=0.5, row sep=-2pt},
        grid=major,
        every axis plot/.append style={line width=0.5pt, mark=*, mark size=1.5pt}
    ]

    \addplot[orange, mark=*] table[x index=0, y index=1] {supp/figures/p_near.txt};
    
    \addplot[blue , mark=*] table[x index=0, y index=1] {supp/figures/p_far.txt};
    
    \addlegendentry{Near}
    \addlegendentry{Far}

    \end{axis}
\end{tikzpicture}

%% file: supp/tables/cifar_expand.tex
\begin{table*}[tbp]
  \centering
  \scriptsize
  \begin{tabularx}{\textwidth}{@{}cX SS R SS R SS  R SS R SS@{}}
    \toprule
    &\multicolumn{1}{X}{} 
    & \multicolumn{2}{c}{\phantom{xii}\textbf{SVHN}} &
    & \multicolumn{2}{c}{\textbf{iSUN}} &
    & \multicolumn{2}{c}{\textbf{Textures}} &
    & \multicolumn{2}{c}{\textbf{Places365}} &
    & \multicolumn{2}{c}{\textbf{Average}}\\
 
\cmidrule(l{6pt}r{0pt}){3-4} \cmidrule(l{0pt}r{0pt}){6-7} \cmidrule(l{0pt}r{0pt}){9-10} \cmidrule(l{0pt}r{0pt}){12-13} \cmidrule(l{0pt}r{0pt}){15-16}

&{\textbf{Method}} 
& {AUC $(\uparrow)$} & {FPR $(\downarrow)$} &
& {AUC $(\uparrow)$} & {FPR $(\downarrow)$} & 
& {AUC $(\uparrow)$} & {FPR $(\downarrow)$} &
& {AUC $(\uparrow)$} & {FPR $(\downarrow)$} &
& {AUC $(\uparrow)$} & {FPR $(\downarrow)$} \\

\midrule

\parbox[t]{2mm}{\multirow{6}{*}{\rotatebox[origin=c]{90}{CIFAR10}}}
&{Energy~\cite{Liu:2020:EOD}} 
& 93.99 & 40.61 &
& 98.07 & 10.07 &
& 86.43 & 56.12 &
& 91.64 & 39.40 &
& 91.18 & 54.18  \\

&{ReAct~\cite{Sun:2021:ODR}} 
& 93.87 & 41.64 &
& 97.72 & 12.72 &
& 92.47 & 43.58 &
& 91.03 & 43.41 &
& 93.77 & 35.31  \\

&{DICE~\cite{Sun:2022:LSO}} 
& 95.90 & 25.99 &
& 99.14 & 4.36 &
& 88.18 & 41.90 &
& 89.13 & 48.59 &
& 93.09 & 30.21  \\

&{LINe~\cite{Ahn:2023:ODL}} 
&97.75& 11.38 &
& 99.01 & 4.90 &
& 95.12 & 23.44 &
& 91.17 & 43.96 &
& 95.75 & 20.88  \\

&{ASH-S~\cite{Djurisic:2023:ESA}} 
& 98.65 & 6.51 &
& 98.90 & 5.17 &
& 95.09 & 24.34 &
& 88.34 & 48.45 &
& 95.25 & 21.12  \\

&{DDCS~\cite{Yuan:2024:DDC}} 
& 97.95 & 9.90 &
& 99.11 & 4.45 &
& 95.96 & 20.16 &
& 91.19 & 42.90 &
& 96.05 & 19.35  \\

&{SCALE~\cite{Xu:2024:STT}} 
& 98.72 & 5.80 &
& \bfseries99.21 & \bfseries3.43 &
& 94.97 & 23.42 &
& 91.74 & 38.69 &
& 96.16 & 17.84  \\

\cmidrule(l{6pt}r{0pt}){2-16}

&{$\combi{S}$} {\bfseries ActSub} (ours) 
& \bfseries99.09 & \bfseries4.39 &
& 99.17 & 3.45 &
& \bfseries96.76 & \bfseries17.38 &
& \bfseries92.47 & \bfseries36.27 &
& \bfseries96.87 & \bfseries15.37  \\

\midrule
\parbox[t]{2mm}{\multirow{6}{*}{\rotatebox[origin=c]{90}{CIFAR100}}}
&{Energy~\cite{Liu:2020:EOD}} 
& 81.85 & 87.46 &
& 78.95 & 74.54 &
& 71.03 & 84.15 &
& 77.72 & 79.20 &
& 77.39 & 81.34  \\

&{ReAct~\cite{Sun:2021:ODR}} 
& 81.41 & 83.81 &
& 86.55 & 65.27 &
& 78.95 & 77.78 &
& 74.04  & 82.65 &
& 80.24  & 77.38  \\

&{DICE~\cite{Sun:2022:LSO}} 
& 88.84 & 54.65 &
& 90.08 & 48.72&
& 76.42 & 65.04 &
& \bfseries77.26 & \bfseries79.58 &
& 83.15 & 62.00  \\

&{LINe~\cite{Ahn:2023:ODL}} 
& 91.90 & 31.10 &
& 94.76 & 24.12 &
& 87.84 & 39.29 &
& 64.18 & 88.41 &
& 84.63 & 45.74  \\

&{ASH-S~\cite{Djurisic:2023:ESA}} 
& 95.76 & 25.02 &
& 91.30 & 46.67 &
& 92.35 & 34.02 &
& 71.62 & 85.86 &
& 87.76 & 47.89  \\

&{DDCS~\cite{Yuan:2024:DDC}} 
& 92.58 & 31.34 &
& \bfseries96.17 & \bfseries18.46 &
& 90.29 & 35.30 &
& 67.91 & 87.11 &
& 86.73 & 43.05  \\

&{SCALE~\cite{Xu:2024:STT}} 
& 96.29 & 22.05 &
& 92.47 & 42.14 &
& 92.34 & 34.20 &
& 72.66 & 85.04 &
& 88.44 & 45.86 \\

\cmidrule(l{6pt}r{0pt}){2-16}

&{$\combi{S}$} {\bfseries ActSub} (ours) 
& \bfseries97.45 & \bfseries13.72 &
& 91.43	& 43.83 &
& \bfseries95.07 & \bfseries23.44 &
& 73.46 & 84.06 &
& \bfseries89.35 & \bfseries41.26  \\ 

    \bottomrule
  \end{tabularx}
  \vspace{-0.5em}
  \caption{\textit{Expanded version of CIFAR results for DenseNet-101, showing each dataset individually} (all in \%). We report results with CIFAR10 \cite{Krizhevsky:2009:LML} and CIFAR100 \cite{Krizhevsky:2009:LML} as ID and SVHN~\cite{Netzer:2011:RDN}, iSUN~\cite{Xu:2015:CSW}, Places365~\cite{Zhou:2018:PMI}, and Textures~\cite{Cimpoi:2014:DTW} as OOD.  
  }
  \label{tab:cifar_exp}
\end{table*}

%% file: supp/tables/openood_expand.tex
\begin{table*}[tbp]
  \centering
  \scriptsize
  \begin{tabularx}{\textwidth}{@{}X R@{} S S S S@{} S S S S@{}}
    \toprule
    {\textbf{Method}}& 
    & {\textbf{NINCO(Near)}} 
    & {\textbf{SSB-Hard(Near)}} 
    & {\textbf{Near-OOD(Avg.)}} &
    & {\textbf{iNaturalist(Far)}} 
    & {\textbf{Textures(Far)}} 
    & {\textbf{OpenImage-O(Far)}} 
    & {\textbf{Far-OOD(Avg.)}}\\
 
\midrule

{MSP~\cite{Hendrycks:2017:BDM}}& 
& {79.95 } / { 56.88}
& {72.09 } / { 74.49}  
& {76.02 } / { 65.68} &  
& {88.41 } / { 43.34}  
& {82.43 } / { 60.87}   
& {84.86 } / { 50.13}    
& {85.23 } / { 51.45}   \\
            
{Energy~\cite{Liu:2020:EOD}}& 
& {79.70 } / { 60.58}  
& {72.08 } / { 76.54}  
& {75.89 } / { 68.56} &  
& {90.63 } / { 31.30}  
& {88.70 } / { 45.77}   
& {89.06 } / { 38.09}    
& {89.47 } / { 38.39}   \\
            
{ReAct~\cite{Sun:2021:ODR}}& 
& {81.73 } / { 55.82}  
& {73.03 } / { 77.55}   
& {77.38 } / { 66.69} &  
& {96.34 } / { 16.72}  
& {92.79 } / { 29.64}   
& {91.87 } / { 32.58}    
& {93.67 } / { 26.31}   \\

{RankFeat~\cite{Koyejo:2022:RFR}}& 
& {55.89 } / { 89.63}  
& {46.08 } / { 94.03}   
& {50.99 } / { 91.83} &  
& {40.06 } / { 94.40}  
& {70.90 } / { 76.84}   
& {50.83 } / { 90.26}    
& {53.93 } / { 87.17}   \\
       
{ViM~\cite{Sun:2022:ODD}}& 
& {78.63 } / { 62.29}  
& {65.54 } / { 80.41}   
& {72.08 } / { 71.35} &  
& {89.56 } / { 30.68}  
& {97.97 } / { 10.51}   
& {90.50 } / { 32.82}    
& {92.68 } / { 24.67}   \\
         
{SHE~\cite{Zhang:2023:ODI}}& 
& {76.49 } / { 69.72}  
& {71.08 } / { 76.30}   
& {73.78 } / { 73.01} &  
& {92.65 } / { 34.06}  
& {93.60 } / { 35.27}   
& {86.52 } / { 55.02}    
& {90.92 } / { 41.45}   \\
       
{GEN~\cite{Liu:2023:PLS}}& 
& {81.70 } / { 54.90}  
& {72.01 } / { 75.73}   
& {76.85 } / { 65.32} &  
& {92.44 } / { 26.10}  
& {87.59 } / { 46.22}   
& {89.26 } / { 34.50}    
& {89.76 } / { 35.61}   \\
            
{ASH-S~\cite{Djurisic:2023:ESA}}& 
& {84.54 } / { 53.26}  
& {74.72 } / { 70.80}   
& {79.63 } / { 62.03} &  
& {97.72 } / { 11.02}  
& {97.87 } / { 10.90}   
& {93.82 } / { 28.60}    
& {96.47 } / { 16.86}   \\

{WeiPer+KLD~\cite{Granz:2024:DWP}}& 
& {85.37 } / { 48.67}  
& {74.73 } / { 74.12}    
& {80.05 } / { 61.39} &  
& {97.49 } / { 13.59}  
& {96.18 } / { 22.17}   
& {92.94 } / { 30.49}    
& {95.54 } / { 22.08}   \\

{SCALE~\cite{Xu:2024:STT}}& 
& {85.37 } / { 51.80}  
& {77.35 } / { 67.72}   
& {81.36 } / { 59.76} &  
& {98.02 } / { 9.51}  
& {97.63 } / { 11.90}   
& {93.95 } / { 28.18}    
& {96.53 } / { 16.53}   \\

\midrule

{$\combi{S}$ {\bfseries ActSub} w/ ReAct}&
& {83.87 } / { 51.01}  
& {71.53 } / { 79.71}   
& {77.70 } / { 65.36} &  
& {97.92 } / { 8.85}  
& {98.13 } / { 8.64}   
& {93.62 } / { 29.23}    
& {96.56 } / { 15.57}   \\

{$\combi{S}$ {\bfseries ActSub} w/ ASH-S}&
& {\bfseries87.42 } / { \bfseries43.36}  
& {80.70 } / { \bfseries61.58}   
& {84.06 } / { \bfseries52.47} &  
& {98.06 } / { 8.48}  
& {98.24 } / { 9.16}   
& {\bfseries94.41 } / { \bfseries25.03}    
& {96.91 } / { \bfseries14.22}   \\

{$\combi{S}$ {\bfseries ActSub} w/ SCALE}&
& {87.35 } / { 43.49}  
& {\bfseries81.14 } / { 61.71}   
& {\bfseries84.24 } / { 52.61} &  
& {\bfseries98.51 } / { \bfseries6.79}  
& {\bfseries98.26 } / { \bfseries8.44}   
& {94.11 } / { 27.54}    
& {\bfseries96.96 } / { 14.29}   \\
    \bottomrule
  \end{tabularx}
  \vspace{-0.5em}
  \caption{\textit{Expanded version of OpenOOD results for ResNet-50 trained on ImageNet-1k, showing each dataset individually} (all in \%). Reported results are in the format {AUC $(\protect\uparrow)$} / {FPR $(\protect\downarrow)$}.}
  \label{tab:openood_exp}
\end{table*}

%% file: main.bbl
\begin{thebibliography}{66}
\providecommand{\natexlab}[1]{#1}
\providecommand{\url}[1]{\texttt{#1}}
\expandafter\ifx\csname urlstyle\endcsname\relax
  \providecommand{\doi}[1]{doi: #1}\else
  \providecommand{\doi}{doi: \begingroup \urlstyle{rm}\Url}\fi

\bibitem[Ahn et~al.(2023)Ahn, Park, and Kim]{Ahn:2023:ODL}
Yong~Hyun Ahn, Gyeong{-}Moon Park, and Seong~Tae Kim.
\newblock {LIN}e: Out-of-distribution detection by leveraging important neurons.
\newblock In \emph{CVPR}, pages 19852--19862, 2023.

\bibitem[Behpour et~al.(2023)Behpour, Doan, Li, He, Gou, and Ren]{Behpour:2023:SEO}
Sima Behpour, Thang~Long Doan, Xin Li, Wenbin He, Liang Gou, and Liu Ren.
\newblock Grad{O}rth: {A} simple yet efficient out-of-distribution detection with orthogonal projection of gradients.
\newblock In \emph{NeurIPS}, 2023.

\bibitem[Bitterwolf et~al.(2023)Bitterwolf, M{\"{u}}ller, and Hein]{Bitterwolf:2023:FIO}
Julian Bitterwolf, Maximilian M{\"{u}}ller, and Matthias Hein.
\newblock In or out? {Fixing} {ImageNet} out-of-distribution detection evaluation.
\newblock In \emph{ICML}, pages 2471--2506, 2023.

\bibitem[Cao and Wu(2022)]{Hao:2022:RCS}
Yun{-}Hao Cao and Jianxin Wu.
\newblock A random {CNN} sees objects: One inductive bias of {CNN} and its applications.
\newblock In \emph{AAAI}, pages 194--202, 2022.

\bibitem[Chan et~al.(2021)Chan, Rottmann, and Gottschalk]{Chan:2021:EMM}
Robin Chan, Matthias Rottmann, and Hanno Gottschalk.
\newblock Entropy maximization and meta classification for out-of-distribution detection in semantic segmentation.
\newblock In \emph{ICCV}, pages 5108--5117, 2021.

\bibitem[Chen et~al.(2023)Chen, Li, Qu, Wang, Wan, and Xiao]{Chen:2023:DGA}
Jinggang Chen, Junjie Li, Xiaoyang Qu, Jianzong Wang, Jiguang Wan, and Jing Xiao.
\newblock {GAIA:} {D}elving into gradient-based attribution abnormality for out-of-distribution detection.
\newblock In \emph{NeurIPS}, 2023.

\bibitem[Cimpoi et~al.(2014)Cimpoi, Maji, Kokkinos, Mohamed, and Vedaldi]{Cimpoi:2014:DTW}
Mircea Cimpoi, Subhransu Maji, Iasonas Kokkinos, Sammy Mohamed, and Andrea Vedaldi.
\newblock Describing textures in the wild.
\newblock In \emph{CVPR}, pages 3606--3613, 2014.

\bibitem[Cook et~al.(2020)Cook, Zare, and Gader]{Cook:2020:ODN}
Matthew Cook, Alina Zare, and Paul~D. Gader.
\newblock Outlier detection through null space analysis of neural networks.
\newblock \emph{arXiv:2007.01263 [cs.LG]}, 2020.

\bibitem[Deng et~al.(2009)Deng, Dong, Socher, Li, Li, and Fei{-}Fei]{Deng:2009:LHI}
Jia Deng, Wei Dong, Richard Socher, Li{-}Jia Li, Kai Li, and Li Fei{-}Fei.
\newblock {ImageNet}: {A} large-scale hierarchical image database.
\newblock In \emph{CVPR}, pages 248--255, 2009.

\bibitem[Dirksen et~al.(2022)Dirksen, Genzel, Jacques, and Stollenwerk]{Dirksen:2022:SCR}
Sjoerd Dirksen, Martin Genzel, Laurent Jacques, and Alexander Stollenwerk.
\newblock The separation capacity of random neural networks.
\newblock \emph{Int. J. Comput. Vision}, 23\penalty0 (309):\penalty0 1--47, 2022.

\bibitem[Djurisic et~al.(2023)Djurisic, Bozanic, Ashok, and Liu]{Djurisic:2023:ESA}
Andrija Djurisic, Nebojsa Bozanic, Arjun Ashok, and Rosanne Liu.
\newblock Extremely simple activation shaping for out-of-distribution detection.
\newblock In \emph{ICLR}, 2023.

\bibitem[Dosovitskiy et~al.(2021)Dosovitskiy, Beyer, Kolesnikov, Weissenborn, Zhai, Unterthiner, Dehghani, Minderer, Heigold, Gelly, Uszkoreit, and Houlsby]{Dosovitskiy:2021:ICLR}
Alexey Dosovitskiy, Lucas Beyer, Alexander Kolesnikov, Dirk Weissenborn, Xiaohua Zhai, Thomas Unterthiner, Mostafa Dehghani, Matthias Minderer, Georg Heigold, Sylvain Gelly, Jakob Uszkoreit, and Neil Houlsby.
\newblock An image is worth 16x16 words: Transformers for image recognition at scale.
\newblock In \emph{ICLR}, 2021.

\bibitem[Fang et~al.(2024)Fang, Tao, Lv, He, Huang, and Yang]{Fang:2024:KPO}
Kun Fang, Qinghua Tao, Kexin Lv, Mingzhen He, Xiaolin Huang, and Jie Yang.
\newblock Kernel {PCA} for out-of-distribution detection.
\newblock In \emph{NeurIPS}, 2024.

\bibitem[Granz et~al.(2024)Granz, Heurich, and Landgraf]{Granz:2024:DWP}
Maximilian Granz, Manuel Heurich, and Tim Landgraf.
\newblock Wei{P}er: {OOD} detection using weight perturbations of class projections.
\newblock In \emph{NeurIPS}, 2024.

\bibitem[Guan et~al.(2023)Guan, Liu, Zheng, Zhou, and Wang]{Guan:2023:RPT}
Xiaoyuan Guan, Zhouwu Liu, Wei{-}Shi Zheng, Yuren Zhou, and Ruixuan Wang.
\newblock Revisit {PCA}-based technique for out-of-distribution detection.
\newblock In \emph{ICCV}, pages 19374--19382, 2023.

\bibitem[He et~al.(2016)He, Zhang, Ren, and Sun]{He:2016:DRL}
Kaiming He, Xiangyu Zhang, Shaoqing Ren, and Jian Sun.
\newblock Deep residual learning for image recognition.
\newblock In \emph{CVPR}, pages 770--778, 2016.

\bibitem[Hendrycks and Gimpel(2017)]{Hendrycks:2017:BDM}
Dan Hendrycks and Kevin Gimpel.
\newblock A baseline for detecting misclassified and out-of-distribution examples in neural networks.
\newblock In \emph{ICLR}, 2017.

\bibitem[Hendrycks et~al.(2019)Hendrycks, Mazeika, and Dietterich]{Hendrycks:2019:DAD}
Dan Hendrycks, Mantas Mazeika, and Thomas~G. Dietterich.
\newblock Deep anomaly detection with outlier exposure.
\newblock In \emph{ICLR}, 2019.

\bibitem[Hendrycks et~al.(2021)Hendrycks, Basart, Mu, Kadavath, Wang, Dorundo, Desai, Zhu, Parajuli, Guo, Song, Steinhardt, and Gilmer]{Hendrycks:2021:MFR}
Dan Hendrycks, Steven Basart, Norman Mu, Saurav Kadavath, Frank Wang, Evan Dorundo, Rahul Desai, Tyler Zhu, Samyak Parajuli, Mike Guo, Dawn Song, Jacob Steinhardt, and Justin Gilmer.
\newblock The many faces of robustness: {A} critical analysis of out-of-distribution generalization.
\newblock In \emph{ICCV}, pages 8320--8329, 2021.

\bibitem[Hendrycks et~al.(2022)Hendrycks, Basart, Mazeika, Zou, Kwon, Mostajabi, Steinhardt, and Song]{Hendrycks:2022:SOD}
Dan Hendrycks, Steven Basart, Mantas Mazeika, Andy Zou, Joseph Kwon, Mohammadreza Mostajabi, Jacob Steinhardt, and Dawn Song.
\newblock Scaling out-of-distribution detection for real-world settings.
\newblock In \emph{ICML}, pages 8759--8773, 2022.

\bibitem[Hofmann et~al.(2024)Hofmann, Schmid, Lehner, Klotz, and Hochreiter]{Hofmann:2024:EHB}
Claus Hofmann, Simon Schmid, Bernhard Lehner, Daniel Klotz, and Sepp Hochreiter.
\newblock Energy-based {H}opfield boosting for out-of-distribution detection.
\newblock In \emph{NeurIPS}, 2024.

\bibitem[Horn et~al.(2018)Horn, Aodha, Song, Cui, Sun, Shepard, Adam, Perona, and Belongie]{Horn:2018:ISC}
Grant~Van Horn, Oisin~Mac Aodha, Yang Song, Yin Cui, Chen Sun, Alexander Shepard, Hartwig Adam, Pietro Perona, and Serge~J. Belongie.
\newblock The i{N}aturalist species classification and detection dataset.
\newblock In \emph{CVPR}, pages 8769--8778, 2018.

\bibitem[Huang et~al.(2017)Huang, Liu, van~der Maaten, and Weinberger]{Huang:2017:DCC}
Gao Huang, Zhuang Liu, Laurens van~der Maaten, and Kilian~Q. Weinberger.
\newblock Densely connected convolutional networks.
\newblock In \emph{CVPR}, pages 770--778, 2017.

\bibitem[Huang and Li(2021)]{HUANG:2021:TSO}
Rui Huang and Yixuan Li.
\newblock {MOS:} {T}owards scaling out-of-distribution detection for large semantic space.
\newblock In \emph{CVPR}, pages 8710--8719, 2021.

\bibitem[Huang et~al.(2021)Huang, Geng, and Li]{Huang:2021:OIG}
Rui Huang, Andrew Geng, and Yixuan Li.
\newblock On the importance of gradients for detecting distributional shifts in the wild.
\newblock In \emph{NeurIPS}, pages 677--689, 2021.

\bibitem[Ioffe and Szegedy(2015)]{Ioffe:2015:ADN}
Sergey Ioffe and Christian Szegedy.
\newblock {Batch Normalization}: Accelerating deep network training by reducing internal covariate shift.
\newblock In \emph{ICML}, pages 448--456, 2015.

\bibitem[Ji and Telgarsky(2020)]{Ji:2020:DCA}
Ziwei Ji and Matus Telgarsky.
\newblock Directional convergence and alignment in deep learning.
\newblock In \emph{NeurIPS}, 2020.

\bibitem[Knuth(1992)]{Knuth:1992:TNN}
Donald~E. Knuth.
\newblock Two notes on notation.
\newblock \emph{The American Mathematical Monthly}, 99\penalty0 (5):\penalty0 403--422, 1992.

\bibitem[Krizhevsky(2009)]{Krizhevsky:2009:LML}
Alex Krizhevsky.
\newblock Learning multiple layers of features from tiny images.
\newblock Technical report, University of Toronto, 2009.

\bibitem[LeCun et~al.(2010)LeCun, Cortes, and Burges]{MNIST}
Yann LeCun, Corinna Cortes, and Chris Burges.
\newblock {MNIST} handwritten digit database, 2010.

\bibitem[Lee et~al.(2018)Lee, Lee, Lee, and Shin]{Lee:2018:SUF}
Kimin Lee, Kibok Lee, Honglak Lee, and Jinwoo Shin.
\newblock A simple unified framework for detecting out-of-distribution samples and adversarial attacks.
\newblock In \emph{NeurIPS}, pages 7167--7177, 2018.

\bibitem[Li et~al.(2024)Li, Xiong, Chen, and Chen]{Li:2024:OSL}
Yixia Li, Boya Xiong, Guanhua Chen, and Yun Chen.
\newblock {SeTAR}: Out-of-distribution detection with selective low-rank approximation.
\newblock In \emph{NeurIPS}, 2024.

\bibitem[Liang et~al.(2018)Liang, Li, and Srikant]{Liang:2018:ERO}
Shiyu Liang, Yixuan Li, and R. Srikant.
\newblock Enhancing the reliability of out-of-distribution image detection in neural networks.
\newblock In \emph{ICLR}, 2018.

\bibitem[Lin et~al.(2021)Lin, Roy, and Li]{Lin:2021:MLO}
Ziqian Lin, Sreya~Dutta Roy, and Yixuan Li.
\newblock {MOOD:} {M}ulti-level out-of-distribution detection.
\newblock In \emph{CVPR}, pages 15313--15323, 2021.

\bibitem[Liu and Qin(2024)]{Liu:2024:FBD}
Litian Liu and Yao Qin.
\newblock Fast decision boundary based out-of-distribution detector.
\newblock In \emph{ICML}, 2024.

\bibitem[Liu et~al.(2020)Liu, Wang, Owens, and Li]{Liu:2020:EOD}
Weitang Liu, Xiaoyun Wang, John~D. Owens, and Yixuan Li.
\newblock Energy-based out-of-distribution detection.
\newblock In \emph{NeurIPS}, 2020.

\bibitem[Liu et~al.(2023)Liu, Lochman, and Zach]{Liu:2023:PLS}
Xixi Liu, Yaroslava Lochman, and Christopher Zach.
\newblock {GEN:} {P}ushing the limits of softmax-based out-of-distribution detection.
\newblock In \emph{CVPR}, pages 23946--23955, 2023.

\bibitem[Lu et~al.(2024)Lu, Gong, Wang, Xue, Yao, and Moore]{LU:2024:LMP}
Haodong Lu, Dong Gong, Shuo Wang, Jason Xue, Lina Yao, and Kristen Moore.
\newblock Learning with mixture of prototypes for out-of-distribution detection.
\newblock In \emph{ICLR}, 2024.

\bibitem[Ming et~al.(2023)Ming, Sun, Dia, and Li]{MING:2023:HEE}
Yifei Ming, Yiyou Sun, Ousmane Dia, and Yixuan Li.
\newblock How to exploit hyperspherical embeddings for out-of-distribution detection?
\newblock In \emph{ICLR}, 2023.

\bibitem[Netzer et~al.()Netzer, Wang, Coates, Bissacco, Wu, and Ng]{Netzer:2011:RDN}
Yuval Netzer, Tao Wang, Adam Coates, Alessandro Bissacco, Bo Wu, and Andrew~Y. Ng.
\newblock Reading digits in natural images with unsupervised feature learning.
\newblock In \emph{NIPS*2011}.

\bibitem[Park et~al.(2023)Park, Jung, and Teoh]{Park:2023:NNG}
Jaewoo Park, Yoon~Gyo Jung, and Andrew Beng~Jin Teoh.
\newblock Nearest neighbor guidance for out-of-distribution detection.
\newblock In \emph{ICCV}, pages 1686--1695, 2023.

\bibitem[Paszke et~al.(2019)Paszke, Gross, Massa, Lerer, Bradbury, Chanan, Killeen, Lin, Gimelshein, Antiga, Desmaison, K{\"{o}}pf, Yang, DeVito, Raison, Tejani, Chilamkurthy, Steiner, Fang, Bai, and Chintala]{Paszke:2019ISH}
Adam Paszke, Sam Gross, Francisco Massa, Adam Lerer, James Bradbury, Gregory Chanan, Trevor Killeen, Zeming Lin, Natalia Gimelshein, Luca Antiga, Alban Desmaison, Andreas K{\"{o}}pf, Edward~Z. Yang, Zachary DeVito, Martin Raison, Alykhan Tejani, Sasank Chilamkurthy, Benoit Steiner, Lu Fang, Junjie Bai, and Soumith Chintala.
\newblock {PyTorch}: An imperative style, high-performance deep learning library.
\newblock In \emph{NeurIPS}, 2019.

\bibitem[Ren et~al.(2021)Ren, Fort, Liu, Roy, Padhy, and Lakshminarayanan]{Ren:2021:ASF}
Jie Ren, Stanislav Fort, Jeremiah~Z. Liu, Abhijit~Guha Roy, Shreyas Padhy, and Balaji Lakshminarayanan.
\newblock A simple fix to {M}ahalanobis distance for improving near-{OOD} detection.
\newblock \emph{arXiv:2106.09022 [cs.LG]}, 2021.

\bibitem[Russakovsky et~al.(2015)Russakovsky, Deng, Su, Krause, Satheesh, Ma, Huang, Karpathy, Khosla, Bernstein, Berg, and Fei{-}Fei]{Russakovsky:2015:INS}
Olga Russakovsky, Jia Deng, Hao Su, Jonathan Krause, Sanjeev Satheesh, Sean Ma, Zhiheng Huang, Andrej Karpathy, Aditya Khosla, Michael~S. Bernstein, Alexander~C. Berg, and Li Fei{-}Fei.
\newblock {ImageNet} large scale visual recognition challenge.
\newblock \emph{Int. J. Comput. Vision}, 115\penalty0 (3):\penalty0 211--252, 2015.

\bibitem[Sandler et~al.(2018)Sandler, Howard, Zhu, Zhmoginov, and Chen]{Sandler:2018:IRL}
Mark Sandler, Andrew~G. Howard, Menglong Zhu, Andrey Zhmoginov, and Liang{-}Chieh Chen.
\newblock {MobileNetV2}: Inverted residuals and linear bottlenecks.
\newblock In \emph{CVPR}, pages 4510--4520, 2018.

\bibitem[Sehwag et~al.(2021)Sehwag, Chiang, and Mittal]{Sehwag:2021:UFS}
Vikash Sehwag, Mung Chiang, and Prateek Mittal.
\newblock {SSD:} {A} unified framework for self-supervised outlier detection.
\newblock In \emph{ICLR}, 2021.

\bibitem[Shapley(1953)]{Shapley:1953:VPG}
L.~S. Shapley.
\newblock \emph{17. A Value for n-Person Games}, pages 307--318.
\newblock Princeton University Press, 1953.

\bibitem[Song et~al.(2022)Song, Sebe, and Wang]{Koyejo:2022:RFR}
Yue Song, Nicu Sebe, and Wei Wang.
\newblock {RankFeat}: Rank-1 feature removal for out-of-distribution detection.
\newblock In \emph{NeurIPS}, 2022.

\bibitem[Strang(2006)]{Strang:2006:LAA}
G. Strang.
\newblock \emph{Linear Algebra and Its Applications}.
\newblock Thomson, Brooks/Cole, 2006.

\bibitem[Sun and Li(2022)]{Sun:2022:LSO}
Yiyou Sun and Yixuan Li.
\newblock {DICE:} {L}everaging sparsification for out-of-distribution detection.
\newblock In \emph{ECCV}, pages 691--708, 2022.

\bibitem[Sun et~al.(2021)Sun, Guo, and Li]{Sun:2021:ODR}
Yiyou Sun, Chuan Guo, and Yixuan Li.
\newblock {ReAct}: Out-of-distribution detection with rectified activations.
\newblock In \emph{NeurIPS}, 2021.

\bibitem[Sun et~al.(2022)Sun, Ming, Zhu, and Li]{Sun:2022:ODD}
Yiyou Sun, Yifei Ming, Xiaojin Zhu, and Yixuan Li.
\newblock Out-of-distribution detection with deep nearest neighbors.
\newblock In \emph{ICML}, pages 20827--20840, 2022.

\bibitem[Ulyanov et~al.(2018)Ulyanov, Vedaldi, and Lempitsky]{Ulyanov:2018:DIP}
Dmitry Ulyanov, Andrea Vedaldi, and Victor~S. Lempitsky.
\newblock Deep image prior.
\newblock In \emph{CVPR}, pages 9446--9454, 2018.

\bibitem[Vaze et~al.(2022)Vaze, Han, Vedaldi, and Zisserman]{Vaze:2022:GCC}
Sagar Vaze, Kai Han, Andrea Vedaldi, and Andrew Zisserman.
\newblock Open-set recognition: {A} good closed-set classifier is all you need.
\newblock In \emph{ICLR}, 2022.

\bibitem[Wang et~al.(2022)Wang, Li, Feng, and Zhang]{Wang:2022:ODV}
Haoqi Wang, Zhizhong Li, Litong Feng, and Wayne Zhang.
\newblock {ViM}: {O}ut-of-distribution with virtual-logit matching.
\newblock In \emph{CVPR}, pages 4911--4920, 2022.

\bibitem[Xiao et~al.(2010)Xiao, Hays, Ehinger, Oliva, and Torralba]{Xiao:2010:LSR}
Jianxiong Xiao, James Hays, Krista~A. Ehinger, Aude Oliva, and Antonio Torralba.
\newblock {SUN} database: Large-scale scene recognition from abbey to zoo.
\newblock In \emph{CVPR}, pages 3485--3492, 2010.

\bibitem[Xu et~al.(2024)Xu, Chen, Franchi, and Yao]{Xu:2024:STT}
Kai Xu, Rongyu Chen, Gianni Franchi, and Angela Yao.
\newblock Scaling for training time and post-hoc out-of-distribution detection enhancement.
\newblock In \emph{ICLR}, 2024.

\bibitem[Xu et~al.(2023)Xu, Lian, Liu, and Tao]{Xu:2023:VRA}
Mingyu Xu, Zheng Lian, Bin Liu, and Jianhua Tao.
\newblock {VRA:} {V}ariational rectified activation for out-of-distribution detection.
\newblock In \emph{NeurIPS}, 2023.

\bibitem[Xu et~al.(2015)Xu, Ehinger, Zhang, Finkelstein, Kulkarni, and Xiao]{Xu:2015:CSW}
Pingmei Xu, Krista~A. Ehinger, Yinda Zhang, Adam Finkelstein, Sanjeev~R. Kulkarni, and Jianxiong Xiao.
\newblock {TurkerGaze}: Crowdsourcing saliency with webcam based eye tracking.
\newblock \emph{arXiv:1504.06755 [cs.CV]}, 2015.

\bibitem[Yang et~al.(2024)Yang, Zhou, Li, and Liu]{Yang:2024:GOD}
Jingkang Yang, Kaiyang Zhou, Yixuan Li, and Ziwei Liu.
\newblock Generalized out-of-distribution detection: {A} survey.
\newblock \emph{Int. J. Comput. Vision}, 132\penalty0 (12):\penalty0 5635--5662, 2024.

\bibitem[Yuan et~al.(2024)Yuan, He, Dong, Han, and Yin]{Yuan:2024:DDC}
Yue Yuan, Rundong He, Yicong Dong, Zhongyi Han, and Yilong Yin.
\newblock Discriminability-driven channel selection for out-of-distribution detection.
\newblock In \emph{CVPR}, pages 26171--26180, 2024.

\bibitem[Zhang et~al.(2023{\natexlab{a}})Zhang, Fu, Chen, Du, Li, Wang, Liu, Han, and Zhang]{Zhang:2023:ODI}
Jinsong Zhang, Qiang Fu, Xu Chen, Lun Du, Zelin Li, Gang Wang, Xiaoguang Liu, Shi Han, and Dongmei Zhang.
\newblock Out-of-distribution detection based on in-distribution data patterns memorization with modern {H}opfield energy.
\newblock In \emph{ICLR}, 2023{\natexlab{a}}.

\bibitem[Zhang et~al.(2023{\natexlab{b}})Zhang, Yang, Wang, Wang, Lin, Zhang, Sun, Du, Zhou, Zhang, Li, Liu, Chen, and Li]{Zhang:2023:EBO}
Jingyang Zhang, Jingkang Yang, Pengyun Wang, Haoqi Wang, Yueqian Lin, Haoran Zhang, Yiyou Sun, Xuefeng Du, Kaiyang Zhou, Wayne Zhang, Yixuan Li, Ziwei Liu, Yiran Chen, and Hai Li.
\newblock {OpenOOD} v1.5: Enhanced benchmark for out-of-distribution detection.
\newblock \emph{arXiv:2306.09301 [cs.LG]}, 2023{\natexlab{b}}.

\bibitem[Zhang and Xiang(2023)]{Zhang:2023:DML}
Zihan Zhang and Xiang Xiang.
\newblock Decoupling {MaxLogit} for out-of-distribution detection.
\newblock In \emph{CVPR}, pages 3388--3397, 2023.

\bibitem[Zhou et~al.(2018)Zhou, Lapedriza, Khosla, Oliva, and Torralba]{Zhou:2018:PMI}
Bolei Zhou, {\`{A}}gata Lapedriza, Aditya Khosla, Aude Oliva, and Antonio Torralba.
\newblock Places: {A} 10 million image database for scene recognition.
\newblock \emph{IEEE T. Pattern Anal. Mach. Intell.}, 40\penalty0 (6):\penalty0 1452--1464, 2018.

\bibitem[Zhu et~al.(2022)Zhu, Chen, Xie, Li, Zhang, Xue, Tian, Zheng, and Chen]{Zhu:2022:BOD}
Yao Zhu, Yuefeng Chen, Chuanlong Xie, Xiaodan Li, Rong Zhang, Hui Xue, Xiang Tian, Bolun Zheng, and Yaowu Chen.
\newblock Boosting out-of-distribution detection with typical features.
\newblock In \emph{NeurIPS}, 2022.

\end{thebibliography}
